%% file: aaai2026.tex
\documentclass[letterpaper]{article} % DO NOT CHANGE THIS
\usepackage{aaai2026}  % DO NOT CHANGE THIS
\usepackage{times}  % DO NOT CHANGE THIS
\usepackage{helvet}  % DO NOT CHANGE THIS
\usepackage{courier}  % DO NOT CHANGE THIS
\usepackage[hyphens]{url}  % DO NOT CHANGE THIS
\usepackage{graphicx} % DO NOT CHANGE THIS
\usepackage{natbib}  % DO NOT CHANGE THIS AND DO NOT ADD ANY OPTIONS TO IT
\usepackage{caption} % DO NOT CHANGE THIS AND DO NOT ADD ANY OPTIONS TO IT
\usepackage{algorithm}
\usepackage{algorithmic}
\usepackage{graphicx}
\usepackage{subfig}
\usepackage{amsmath} 
\usepackage{multirow}
\usepackage{booktabs}

\usepackage{newfloat}
\usepackage{listings}
\DeclareCaptionStyle{ruled}{labelfont=normalfont,labelsep=colon,strut=off} % DO NOT CHANGE THIS
\floatstyle{ruled}
\newfloat{listing}{tb}{lst}{}
\floatname{listing}{Listing}
\title{FlowletFormer: Network Behavioral Semantic Aware Pre-training Model for Traffic Classification}
\author{
    Liming Liu \equalcontrib \textsuperscript{\rm 1},
    Ruoyu Li \equalcontrib \textsuperscript{\rm 2},
    Qing Li \textsuperscript{\rm 3},
    Meijia Hou \textsuperscript{\rm 4},
    Yong Jiang \textsuperscript{\rm 1} \textsuperscript{\rm 3},
    Mingwei Xu \textsuperscript{\rm 5}
}
\affiliations{
    \textsuperscript{\rm 1} Tsinghua Shenzhen International Graduate School
    \textsuperscript{\rm 2} Shenzhen University\\
    \textsuperscript{\rm 3} Peng Cheng Laboratory
    \textsuperscript{\rm 4} Zhongguancun Laboratory
    \textsuperscript{\rm 5} Tsinghua University

}

\usepackage{bibentry}
\begin{document}

\maketitle

\begin{abstract}
Network traffic classification using pre-training models has shown promising results, but existing methods struggle to capture packet structural characteristics, flow-level behaviors, hierarchical protocol semantics, and inter-packet contextual relationships. To address these challenges, we propose FlowletFormer, a BERT-based pre-training model specifically designed for network traffic analysis. FlowletFormer introduces a Coherent Behavior-Aware Traffic Representation Model for segmenting traffic into semantically meaningful units, a Protocol Stack Alignment-Based Embedding Layer to capture multilayer protocol semantics, and Field-Specific and Context-Aware Pretraining Tasks to enhance both inter-packet and inter-flow learning. Experimental results demonstrate that FlowletFormer significantly outperforms existing methods in the effectiveness of traffic representation, classification accuracy, and few-shot learning capability. Moreover, by effectively integrating domain-specific network knowledge, FlowletFormer shows better comprehension of the principles of network transmission (e.g., stateful connections of TCP), providing a more robust and trustworthy framework for traffic analysis.
\end{abstract}


\section{Introduction}

% 整篇文章的核心：
% 当前流量分类预训练模型机械照搬 NLP 技术，忽略了网络数据包的独特性以及网络领域的专家知识。因此，我们提出了自己的方法。

Network traffic refers to data transmitted across networks, including the exchange of packets and other forms of communication between devices. It comprises payload and metadata that provide key insights into network behavior. Monitoring and analyzing traffic is essential for both network management and security \cite{PapadogiannakiI21,TangZeroWall2020}, enabling network operators to effectively tailor resource allocation, guarantee quality of service, and detect malicious activities \cite{requetqosGuttermanGAWWKZ19,towardHuGCLZW23,CJSMaoSVMA19}.

Traditional rule-based methods for traffic classification \cite{Roesch99snort} struggle with complex, encrypted traffic in modern networks \cite{TaylorSCM16}. The researchers then turned to machine learning (ML), which proved effective in traffic classification \cite{AlAdaptive, PanchenkoWebsite},  although it depends on manual feature engineering, requiring expert knowledge and considerable effort. 
Deep Learning (DL) methods address these challenges by automatically extracting features from raw traffic, capturing complex patterns \cite{SirinamIJW18DF, LiuHXCL19FSNET, ShenZZXD21GraphDapp, SchusterST17Beauty}. However, they require large-scale labeled datasets, which are difficult to obtain \cite{DeeplearningRezaeiL19,OptimizingShenLZXDG20,NetworkAouediPHK22}. 

Recently, pre-training methods \cite{HeYC20PERT,ZhaoYATC, Linetbert, ZhouTrafficFormer} have emerged and exhibited superior performance in traffic classification tasks. These methods involve pre-training model on a large volume of unlabeled data to learn general representations, which can subsequently be fine-tuned for specific traffic classification tasks using smaller labeled datasets. 
Approaches like PERT \cite{HeYC20PERT}, ET-BERT \cite{Linetbert}, and Trafficformer \cite{ZhouTrafficFormer} employing BERT models have demonstrated promising results. 

Despite achieving promising accuracy on given datasets, existing pre-training models for traffic classification still exhibit significant limitations.

\textbf{First}, traffic corpora construction often mechanically adopts NLP techniques, such as  4-hex Bigram Encoding and subword tokenization, overlooking packet structure and field semantics.  4-hex Bigram Encoding represents traffic as ``words'', which fail to reflect critical 1- to 2-hex fields (e.g., IP version, TTL). Subword tokenization is used, but fewer than 1\% of ``words'' are further segmented.

% \textbf{First}, in constructing corpora for network traffic, existing approaches often mechanically adopt NLP techniques while neglecting structural characteristics of packets and semantic coherence within flows. For instance, methods such as ET-BERT \cite{Linetbert} encode traffic into words'' using 4-hex bigrams, which fail to effectively reflect critical 1- to 2-hex fields (e.g., IP version, time to live) in the vector space. 
% Although pre-training methods employ subword tokenization for fine-grained features \cite{ZhouTrafficFormer}, Figure \ref{subfig:wordpie} shows that fewer than 1\% of words'' are segmented, with most remaining intact as 4-hex bigrams.
% Additionally, in ET-BERT, we observe that 65\% of the sentences'' contain only a single packet, as shown in Figure \ref{subfig:cdf}, severely limiting the model’s ability to learn behavioral patterns formed by sequential packets within a flow (e.g., web browsing, file transfers).

\textbf{Second}, existing methods fail to model the hierarchical semantics of network protocols, treating packets as flat sequences and ignoring structural differences across layers. For example, the first two bytes at the IP layer and the first two bytes at the TCP layer may have identical HEX values but represent entirely different fields.

\textbf{Third}, existing pretraining tasks are limited in their ability to capture flow-level semantics. Due to 65\% of ``sentence'' containing only a single packet, the SBP tasks in ET-BERT largely degrade into intra-packet association predictions while only 20\% support complete traffic behavioral pattern learning. % Moreover, intra-packet predictions suffer from ambiguity and overfitting (Fig.\ref{subfig:losscurves}) due to weak and inconsistent semantic. As a result, models struggle to learn critical knowledge such as 5-tuple consistency, TCP handshake order, and sequence number semantics.

% \textbf{Third}, current pre-training tasks aim to help models learn general patterns in network traffic, with MLM tasks focusing on intra-packet structural features and SBP tasks targeting traffic patterns. However, since about 65\% of sentences contain only a single packet, the effectiveness of SBP tasks is significantly diminished. Approximately 40\% of SBP tasks effectively become intra-packet association predictions, and only about 20\% fully support learning complete traffic patterns.
% In Addition, intra-packet association prediction task exhibit ambiguity and uncertainty. As shown in Figure \ref{subfig:losscurves}, compared to inter-packet prediction tasks (used as a contrary in our design), intra-packet tasks are more prone to overfitting. Because intra-packet patterns tend to be vague, diverse, and lack strong, consistent semantic anchors.
% As a result, models fail to understand critical domain knowledge, such as 5-tuple consistency within a flow, the sequence of packets in a TCP three-way handshake, and the selection mechanism of TCP sequence numbers. The design and implementation details of the preliminary experiments can be found in \ref{Appendix:preliminary}.

As a consequence of these limitations, existing approaches are hard to learn network contextual information and general patterns in network traffic. To address these challenges, we propose FlowletFormer, a BERT-based pre-training model for network traffic analysis. Specifically, we make the following contributions:

1) To overcome the limitations in existing traffic representation, we propose a coherent behavioral unit, \textbf{Flowlet}, as our traffic representation. Flowlets aggregate packets within a logical interaction. To encode them effectively, we further introduce \textbf{Field Tokenization}, which converts each flowlet into semantically meaningful tokens based on protocol header fields.

2) To address the overlook of the hierarchical structure of network protocols in existing models, we propose a \textbf{Protocol Stack Alignment-Based Embedding Layer} that aligns input tokens with the hierarchical structure of protocols. By encoding multi-layer semantics, our method enables the model to distinguish fields across protocol boundaries and better capture protocol-specific behaviors.

3) To bridge the gaps in pretraining tasks, we introduce two novel pretraining tasks. The \textbf{Masked Field Model} enhances field-level semantic understanding by predicting selectively masked critical protocol fields. The \textbf{Flowlet Prediction Task} captures logical interactions by modeling relations between Flowlets, such as HTTP requests and disconnections.
By incorporating both intra-packet and inter-packet context, our method captures general traffic patterns.

We evaluate FlowletFormer on 8 fine-tuning datasets, achieving state-of-the-art performance on 7 of them, with over 5\% F1 improvement on 5 datasets. 
To assess the capability of the pre-trained model\footnote{\label{ft:pretrained}That is, the model after pretraining but before fine-tuning.} in understanding the underlying principles of network transmission and protocol header semantics, we novelly propose several \textbf{field understanding tasks} and \textbf{word analogy similarity analyses}.
Our code is available in supplementary material.
% The results show that our method outperforms baselines by demonstrating superior comprehension of protocol-specific mechanisms and traffic interaction logic, demonstrating a more trustworthy framework for traffic analysis.

\begin{figure*}[htbp]
\begin{center}
\centerline{\includegraphics[width=1.4\columnwidth]{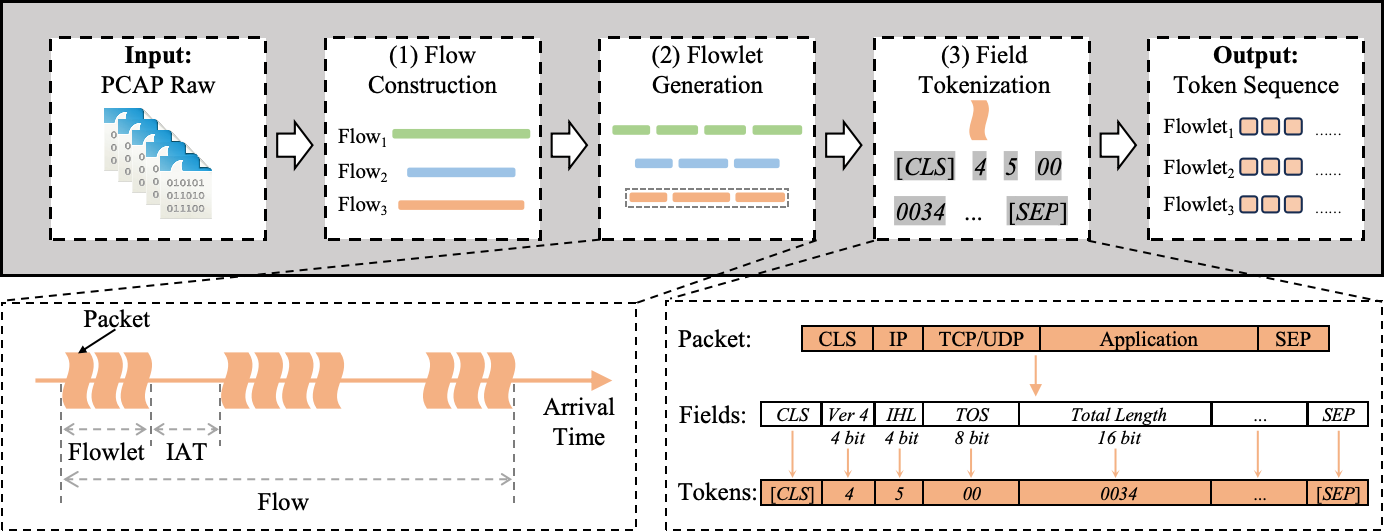}}
\caption{Flowlet and Field Tokenization.}
\label{flowlet}
\end{center}
\end{figure*}

\section{Related Work}
\subsection{Traffic Classification}
Traffic classification methods have evolved significantly in the past decade, driven by the increasing complexity of network traffic and the need for efficient network management. Early approaches to traffic classification focused primarily on the statistical analysis of packet headers and flow characteristics. For example, early work \cite{Roesch99snort, trafficclasszuev2005} utilized packet-based, flow-based features or rule matching, such as packet size and inter-arrival times. However, it becomes ineffective in encrypted traffic where observable patterns are largely concealed.

Early ML-based methods, such as those proposed in \cite{TaylorSCM16, AlAdaptive,PanchenkoWebsite, Sommerml10}, explored classifiers such as decision trees, random forests, and SVMs to classify traffic. These approaches leveraged features including statistical summaries of flow-level metrics and protocol-specific characteristics. Feature engineering requires significant domain expertise and is time-consuming to engineer.

In recent years, deep learning techniques have been increasingly applied to traffic classification, offering superior performance due to their ability to learn complex, high-dimensional representations of network traffic. Lotfollahi \cite{deeppacketlotf20} introduced a DNN-based approach that directly uses raw packet data, bypassing feature extraction. This end-to-end approach significantly improved accuracy over other methods. CNNs, RNNs, and GNNs have been widely applied to traffic classification tasks\cite{SirinamIJW18DF, LiuHXCL19FSNET, ShenZZXD21GraphDapp, SchusterST17Beauty, cnnzhang2020}. However, their high performance often depends on large-scale labeled datasets, which are expensive to obtain in practice.

\subsection{Pre-training Methods}
Pre-training methods have brought breakthroughs to both NLP and CV by enabling models to learn from large-scale unlabeled data. Early models like ELMo \cite{emlo2018} used bidirectional LSTMs to capture contextual information. Later, Transformer-based architectures \cite{vaswani2017attention} such as BERT \cite{bertdevlin2019} and GPT \cite{radford2018improving} popularized pre-training through tasks like Masked Language Modeling. RoBERTa \cite{robertaliu2018} further improved BERT by removing the Next Sentence Prediction objective and increasing training scale, while ALBERT \cite{albertlan2020} reduced model size through parameter sharing.

Due to its powerful sequence modeling capabilities, the transformer architecture has also been widely applied to network traffic classification tasks \cite{HeYC20PERT,ZhaoYATC, Linetbert, ZhouTrafficFormer}. Recent works have adapted Transformer-based pre-training to network traffic. 
PERT \cite{HeYC20PERT} directly feeds raw packets into a Transformer, while YaTC \cite{ZhaoYATC} converts flows into structured images for masked autoencoder training. 
BERT and TrafficFormer \cite{Linetbert, ZhouTrafficFormer} segment flow into BURST by changes in the direction of packets, which is considered to represent transmission patterns from the application layer. Then each packet in BURST is serialized into a hexadecimal string and represents by 4-hex Bigram Encoding (e.g., \colorbox{lightgray}{\texttt{0x45}} \colorbox{lightgray}{\texttt{0x00}} \colorbox{lightgray}{\texttt{0x00}} \colorbox{lightgray}{\texttt{0x34 ...}} into \colorbox{lightgray}{\texttt{4500}}, \colorbox{lightgray}{\texttt{0000}}, \colorbox{lightgray}{\texttt{0034}}, ...). Finally, they apply Subword Tokenization \cite{subwordChungCB16,SennrichHB16aBPE,subwordLuongM16} over the bigram tokens to build a fixed‐size vocabulary and split tokens (e.g., \colorbox{lightgray}{\texttt{1df8}} into \colorbox{lightgray}{\texttt{1df}} and \colorbox{lightgray}{\texttt{\#8}}). The resulting tokens are then used for both pre-training and downstream tasks.

However, these designs based on BURST, 4-hex Bigram Encoding and Subword Tokenization do not fully align with the common patterns of network traffic or the semantic features within packets, which limits the model's understanding of network traffic.

\section{FlowletFormer}
FlowletFormer introduces a novel framework that enables the model to capture fine-grained network behaviors and hierarchical semantics in traffic.

\subsection{Flowlet and Field Tokenization}
Current pre‐training models repurpose NLP representation and tokenization for network traffic, ignoring their distinct structure and semantics.

To overcome this issue, we propose \textbf{Flowlet and Field Tokenization} as shown in Figure \ref{flowlet}.
Flowlet is a fine-grained flow segment and coherent behavior unit that groups packets from the same logical interaction (e.g., an HTTP request-response or a media stream).
Flowlet also emphasizes Temporal Correlation, ensuring that packets transmitted within the same time frame are analyzed together.
And Field Tokenization transforms each flowlet into discrete tokens based on protocol header fields, preserving field-level semantics for model input.
Flowlet and Field Tokenization consists of 3 steps: \textbf{Flow Construction}, \textbf{Flowlet Generation} and  \textbf{Field Tokenization}.

\textbf{Flow Construction.} 
Raw traffic is unordered and mixes multiple protocols, making pattern learning difficult. To impose semantic structure, we group packets by identical five‐tuples and follow the relevant RFCs \cite{ippostel1981internet, tcprfc9293, udppostel1980rfc0768, icmppostel1981internet} to form flows.

\textbf{Flowlet Generation.} 
Consider a flow $\boldsymbol{F}$ consisting of a sequence of $n$ packets, denoted as $\boldsymbol{F} = \{pkt_1, pkt_2, \ldots, pkt_n\}$. Each packet $pkt_i$ has an arrival timestamp $\tau_i$. The objective of Flowlet Generation is to segment this flow into multiple flowlets based on Inter-Arrival Time (IAT) between consecutive packets.

Let us define the IAT between consecutive packets as $t_i = \tau_i - \tau_{i-1}$ for $i \in {2, 3, \ldots, n}$. We introduce a dynamic threshold \( \theta_i \) to determine flowlet boundaries, which is adaptively adjusted based on the historical IATs. Let \( W_i \) denote the IAT window up to the \( i \)-th packet. The threshold is calculated as:
\[
\theta_i = \frac{1}{|W_i|} \sum_{t \in W_i} t
\]

For each flowlet \(\mathcal{F}_j= \{ \text{pkt}_a, \text{pkt}_{a+1}, \dots, \text{pkt}_b \} \), the inter-arrival times within the flowlet satisfy:
\[
t_i \leq \theta_{i-1}, \quad \forall i \in \{a+1, \dots, b\}.
\]
If \( \text{pkt}_b \) is the last packet of flowlet \( F_j \), and \( \text{pkt}_{b+1} \) is the first packet of flowlet \( \mathcal{F}_{j+1}\), then:
\[
t_{b+1} > \theta_b.
\]
The algorithm starts by constructing the first flowlet from the first packet and then processes each packet sequentially. When \( i > 3 \) and the current IAT \( t_i \) exceeds the threshold \( \theta_{i-1} \), a new flowlet boundary is established. Otherwise, the packet is added to the current flowlet. The algorithm continuously updates the window \( W_i \) and adjusts the threshold accordingly to adapt to changing network conditions. More details about flowlet are provided in the Appendix B.
% 时空复杂度
% The time complexity of the algorithm is \( O(n) \), and the space complexity is \( O(w + m) \), where \( w \) is the window size and \( m \) is the number of flowlets generated. This adaptive segmentation method provides an effective foundation for traffic engineering and load balancing in dynamic network environments.

\begin{figure*}[htbp]
\begin{center}
\centerline{\includegraphics[width=0.8\textwidth]{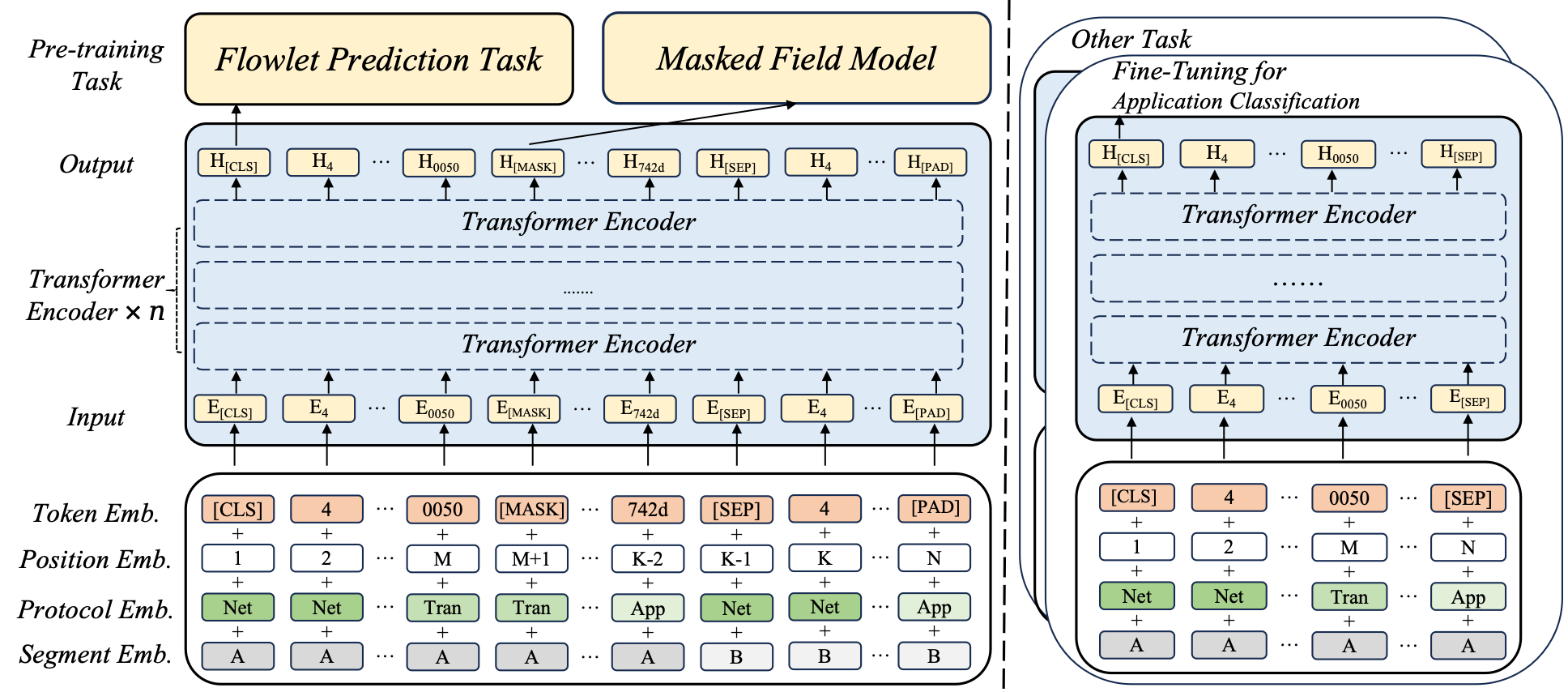}}
\caption{The flowchart of the FlowletFormer.}
\label{flowletformer}
\end{center}
\end{figure*}

\textbf{Field Tokenization.} We transform Flowlets into tokens that can be input into the model. For each packet in the flowlet, we first extract the raw hexadecimal sequences (e.g. \colorbox{lightgray}{\texttt{0x45000034...}}). Field tokenization is based on the length of protocol header fields, encoding the sequence into multiple hexadecimal tokens (e.g. \colorbox{lightgray}{\texttt{0x45000034...}} into \colorbox{lightgray}{\texttt{4}} \colorbox{lightgray}{\texttt{5}} \colorbox{lightgray}{\texttt{00}} \colorbox{lightgray}{\texttt{0034}} ...). 
For fields longer than 2 bytes and payload, we split them into multiple 4-digit hexadecimal tokens to ensure uniformity and consistency in the model input format. 

In this work, we use word-based tokenization \cite{mielke2021between} rather than subword tokenization \cite{subwordChungCB16,SennrichHB16aBPE,subwordLuongM16}, such as BPE \cite{SennrichHB16aBPE, bpefirst} or WordPiece \cite{wu2016googlesneuralmachinetranslation}. The motivation behind this is that, we treat protocol header fields as the morpheme (smallest semantic units) in traffic, similar to individual characters in Chinese. In such languages, each character is a complete and indivisible unit of meaning. Likewise, each protocol field inherently carries distinct and atomic semantics, and should not be further split or processed using subword tokenization methods.

The maximum vocabulary size, denoted as $|V|$, is 65,812. This includes all possible tokens: 1-hex tokens (16 values), 2-hex tokens (256 values), 4-hex tokens (65,536 values), and five special tokens ([CLS], [SEP], [PAD], [MASK], [UNK]).% [CLS] marks the beginning of the sequence and aggregates flowlet-level information. [SEP] indicates the boundary between sequences. [PAD] is used to pad sequences to the same length, ensuring that all input sequences have a consistent size. [MASK] is used to mark parts of the sequence that the model needs to predict, while [UNK] represents tokens that are not found in the dictionary.

% Overall, this representation model and tokenization help FlowletFormer focusing on structural characteristics of network packets and semantic coherence within flows.

\subsection{Model Architecture}
FlowletFormer adopts the BERT-base model architecture \cite{bertdevlin2019}, which consists of two modules: an Embedding Module and a Transformer Encoder Module, as illustrated in Figure \ref{flowletformer}.

\textbf{Embedding Module.} Most existing pre-training models for traffic classification directly adopt the embedding module design for NLP, including token, position, and segment embedding. However, directly using these embeddings may overlook the unique characteristics of traffic. Unlike natural language, traffic has a layered protocol structure, different forms of alignment and distribution.

Thus, we introduce a \textbf{Protocol Stack Alignment-Based Embedding Layer} into the existing embedding module. This embedding layer is specifically designed for traffic data and explicitly encodes the protocol layer associated with each token. Specifically, this embedding distinguishes between the network layer, transport layer, and application layer based on the TCP/IP model \cite{Jamescomputernetwork}, and assigns each token an embedding corresponding to its protocol layer.

This design captures the semantic differences between different protocol layers. The model can not only process tokens based on their positions and sequential order, but also understand their functional roles within the protocol layer. This enables a hierarchical representation of traffic.

Finally, the embedding dimension is set to $D$ = 768 and the input are calculated by the sum of each embedding layer:
\begin{equation}
\mathbf{E}_{\text{input}} = \mathbf{E}_{\text{token}} + \mathbf{E}_{\text{position}} + \mathbf{E}_{\text{segment}} + \mathbf{E}_{\text{protocol}}
\end{equation}

\textbf{Transformer Encoder Module.} 
FlowletFormer is built on the BERT-Base architecture and contains 12 transformer encoder layers \cite{vaswani2017attention}, each with 12 multi-head self-attention heads and a position-wise feedforward network. Residual connections and layer normalization throughout the model ensure stable training and faster convergence. The total number of parameters is approximately 110 million. The number of input tokens is 512, and the dimension of each input token is 768.

\begin{table}[h]
\begin{tabular}{c|l}
\toprule
\textbf{Protocol} & \textbf{Key Fields} \\
\hline
IP      & Version, Total Length, Protocol, IP Address  \\
\hline
\multirow{2}{*}{TCP}     & Port Number, Sequence Number, Flag, \\
        & Acknowledgment Number, Window Size \\
\hline
UDP     & Port Number \\
\hline
ICMP    & Type, Code  \\
\bottomrule
\end{tabular}
\caption{Key fields in common protocol.}
\label{keyfields}
\end{table}

\begin{table*}[htbp!]
    \small
    \centering
    \tabcolsep=0.5mm
    \begin{tabular}{c|cccc|cccc|cccc|cccc}
    \toprule
    Dataset & \multicolumn{4}{c|}{ISCX-VPN(Service)} & \multicolumn{4}{c|}{ ISCX-Tor2016} & \multicolumn{4}{c|}{ISCX-VPN(APP)} & \multicolumn{4}{c}{CSTNET-TLS}  \\
    \cline{1-17}
    Metric & AC & PR & RC & F1 & AC & PR & RC & F1 & AC & PR & RC & F1 & AC & PR & RC & F1\\
    \hline
    AppScanner      & 0.8681 & 0.8710 & 0.8435 & 0.8546 & 0.9075 & 0.7728 & 0.8033 & 0.7848 & 0.7945 & 0.6950 & 0.6975 & 0.6874 & 0.7441 & 0.7232 & 0.6963 & 0.7023 \\
    BIND            & 0.8345 & 0.7769 & 0.7714 & 0.7699 & 0.9010 & 0.8582 & 0.8354 & 0.8439 & 0.6951 & 0.6266 & 0.5415 & 0.5609 & 0.4710 & 0.4315 & 0.4226 & 0.4189 \\
    CUMUL           & 0.7099 & 0.6959 & 0.6893 & 0.6884 & 0.7725 & 0.6463 & 0.6443 & 0.6401 & 0.5480 & 0.4839 & 0.4615 & 0.4554 & 0.5921 & 0.5528 & 0.5604 & 0.5493 \\
    DF              & 0.5018 & 0.4664 & 0.4773 & 0.3934 & 0.7401 & 0.5918 & 0.5611 & 0.5492 & 0.4935 & 0.2449 & 0.2592 & 0.2289 & 0.5729 & 0.5398 & 0.5144 & 0.4933 \\
    FSNet           & 0.9087 & 0.9051 & 0.9054 & 0.9051 & 0.6967 & 0.6159 & 0.6061 & 0.6028 & 0.6316 & 0.4899 & 0.4833 & 0.4677 & 0.7814 & 0.7670 & 0.7316 & 0.7311 \\
    GraphDApp       & 0.7500 & 0.7311 & 0.7678 & 0.7429 & 0.7949 & 0.6391 & 0.6410 & 0.6383 & 0.5703 & 0.5108 & 0.4872 & 0.4853 & 0.7281 & 0.6964 & 0.6909 & 0.6890 \\
    Beauty          & 0.6416 & 0.5769 & 0.5842 & 0.5387 & 0.3746 & 0.2691 & 0.2767 & 0.2251 & 0.6169 & 0.3333 & 0.3139 & 0.2964 & 0.2944 & 0.3219 & 0.2513 & 0.2324 \\
    ET-BERT         & 0.8467 & 0.8496 & 0.8651 & 0.8393 & 0.8123 & 0.7249 & 0.8276 & 0.7453 & 0.7964 & 0.7332 & 0.7013 & 0.7066 & 0.7993 & 0.7832 & 0.7689 & 0.7700 \\
    YaTC            & 0.9010 & 0.8877 & 0.8800 & 0.8821 & 0.9175 & 0.7725 & 0.7333 & 0.7405 & 0.8214 & 0.7443 & 0.7265 & 0.7254 & 0.8391 & 0.8364 & 0.8101 & 0.8140 \\
    TrafficFormer   & 0.8533 & 0.8445 & 0.8348 & 0.8279 & 0.8669 & 0.7545 & 0.7460 & 0.7472 & 0.7751 & 0.7488 & 0.6846 & 0.6962 & 0.7982 & 0.7883 & 0.7736 & 0.7704 \\
    \hline
    FlowletFormer & \textbf{0.9400} & \textbf{0.9471} & \textbf{0.9277} & \textbf{0.9364} & \textbf{0.9215} & \textbf{0.9263} & \textbf{0.9043} & \textbf{0.9116} & \textbf{0.8480} & \textbf{0.8153} & \textbf{0.7641} & \textbf{0.7712} & \textbf{0.8605} & \textbf{0.8578} & \textbf{0.8445} & \textbf{0.8473} \\
    \bottomrule
    \end{tabular}
\caption{Comparison results on ISCXVPN2016, ISCX-Tor2016, and CSTNET-TLS 1.3 datasets.}
\label{tab:metrics1}
\end{table*}

\subsection{Pre-training Method}

We introduce the two novel pretraining tasks separately, followed by the overall loss function.

\textbf{Masked Field Model.} The masked modeling task randomly masks tokens and predicts the masked. Previous studies typically use this task to learn context and dependencies. However, in network traffic, the context and dependencies carried by different tokens vary in importance. Random masking may not fully capture the structural characteristics. 

Thus, we introduce the Masked Field Model.
During pre-training, 15\% of the tokens in the input sequence are masked. \textbf{Half of these masked tokens are randomly selected from the key field tokens} mentioned in Table \ref{keyfields}, while the other half are randomly selected from the remaining tokens. For the masked tokens, we replace them with the token [MASK], a random token, or leave them unchanged with probabilities of 80 \%, 10 \%, and 10 \%, respectively.

For masked tokens, FlowletFormer must predict the token based on the context during pre-training. The loss function used is the cross-entropy loss, as shown in Equation \ref{equationmask}.
\begin{equation}
\label{equationmask}
\mathcal{L}_{\text{MFM}} = - \sum_{i \in \mathcal{M}_{\text{field}} \cup \mathcal{M}_{\text{random}}} m_i \log(\hat{m_i})
\end{equation}

\textbf{Flowlet Prediction Task.} Flowlet is generated based on the IAT between packets, which makes Flowlet more aligned with real network interactions, providing a better representation of network behavior and traffic patterns. For example, in a file download activity, a flow may represent the entire process of downloading the file, while \textbf{each Flowlet reflects specific behavior phases within the network interaction}, such as the request phase, download phase, and disconnection phase.

To better capture the diverse patterns in traffic, we introduce the Flowlet Prediction Task to predict the relationships between Flowlets. During pre-training, we sample a pair of flowlets (\( \mathcal{F}_A\), \( \mathcal{F}_B\)) and form the training instance. The pair is then drawn uniformly from three scenarios: \( \mathcal{F}_B\) is either the immediate successor of \( \mathcal{F}_A\) in the same flow (Ordered), the immediate predecessor (Swapped), or from a different flow. This design forces the model to learn intra‐flow continuity, reverse‐order dynamics, and separation of unrelated flowlets.

% Class 1 and Class 2 help the model understand the temporal and behavioral patterns of flows by capturing continuity and the order of behavior within the same communication process. Class 1 emphasizes the temporality and contextual relevance within the same flow, while Class 2 focuses on capturing the directional changes in the flow (e.g., request and response). Class 3 focuses on distinguishing between different flows, ensuring that unrelated flows are not confused, thereby enabling accurate differentiation of distinct network interactions. 
Unlike tasks based on individual packet or burst \cite{Linetbert,ZhouTrafficFormer}, this task shifts the focus from individual packets to the relationships between behaviorally coherent Flowlets. Its goal is to capture the temporal and behavioral patterns of network traffic beyond the low-level semantics of individual packets.

Finally, the flowlet prediction task utilizes cross-entropy as the loss function, as shown in Equation \ref{equationflowlet}.
\begin{equation}
\label{equationflowlet}
\mathcal{L}_{\text{FPT}} = - \sum_{i=1}^{N} y_i \log(\hat{y_i})
\end{equation}
Overall, the final pre-training objective is the sum of the two losses mentioned above, defined as:
\begin{equation}
\label{overall}
\mathcal{L} = \mathcal{L}_{\text{MFM}} + \mathcal{L}_{\text{FPT}}
\end{equation}
\subsection{Fine-tuning Method}
FlowletFormer acquires generalizable knowledge during pre-training, capturing diverse patterns in traffic rather than being restricted to a specific task. This broad understanding enhances its transferability to various downstream applications. During fine-tuning, the pre-training knowledge is adapted to downstream traffic classification tasks.

\begin{table*}[htbp]
    \small
    \centering
    \tabcolsep=0.5mm
    \begin{tabular}{c|cccc|cccc|cccc|cccc}
    \toprule
    Dataset & \multicolumn{4}{c|}{Browser} & \multicolumn{4}{c|}{USTC-TFC} & \multicolumn{4}{c|}{CIC-IDS2017} & \multicolumn{4}{c}{CIC-IoT2022}  \\
    \cline{1-17}
    Metric & AC & PR & RC & F1 & AC & PR & RC & F1 & AC & PR & RC & F1 & AC & PR & RC & F1\\
    \hline
    AppScanner      & 0.5825 & 0.5836 & 0.6023 & 0.5733 & 0.8585 & 0.9108 & 0.9034 & 0.8976 & 0.8712 & 0.8820 & 0.8697 & 0.8630  & 0.8591 & 0.8858 & 0.7996 & 0.8288 \\
    BIND            & 0.5842 & 0.5875 & 0.5704 & 0.5738 & 0.7945 & 0.7811 & 0.7061 & 0.7115 & 0.9114 & 0.9252 & 0.8652 & 0.8788 & 0.7349 & 0.6754 & 0.6387 & 0.6435 \\
    CUMUL           & 0.5253 & 0.5226 & 0.5202 & 0.5207 & 0.7173 & 0.5063 & 0.5812 & 0.5183 & 0.8478 & 0.6867 & 0.7160 & 0.6951 & 0.7019 & 0.6746 & 0.7029 & 0.6687 \\
    DF              & 0.2450 & 0.1966 & 0.2454 & 0.1627 & 0.6452 & 0.4019 & 0.3685 & 0.3059 & 0.6672 & 0.6756 & 0.6348 & 0.5940 & 0.2746 & 0.2140 & 0.1870 & 0.1647 \\
    FSNet           & 0.6405 & 0.6424 & 0.6491 & 0.6410 & 0.7558 & 0.8167 & 0.8407 & 0.8042 & 0.8436 & 0.8198 & 0.8106 & 0.8144 & 0.8077 & 0.8250 & 0.8333 & 0.7804 \\
    GraphDApp       & 0.4844 & 0.5005 & 0.4969 & 0.4912 & 0.8750 & 0.8446 & 0.8501 & 0.8249 & 0.8750 & 0.8989 & 0.8515 & 0.8647 & 0.7370 & 0.6721 & 0.7006 & 0.6767 \\
    Beauty          & 0.2250 & 0.1393 & 0.2364 & 0.1040 & 0.6682 & 0.4448 & 0.4369 & 0.3796 & 0.6641 & 0.7720 & 0.6482 & 0.6567 & 0.1356 & 0.0349 & 0.0764 & 0.0296 \\
    ET-BERT         & 0.4700 & 0.4861 & 0.4700 & 0.3439 & 0.9663 & 0.9711 & 0.9663 & 0.9666 & 0.8950 & 0.9000 & 0.8950 & 0.8911 & 0.8603 & 0.8297 & 0.8255 & 0.8244 \\
    YaTC            & 0.5276 & 0.5348 & 0.5278 & 0.5154 & 0.9712 & 0.9732 & 0.9712 & 0.9707 & 0.9083 &	0.9332 & 0.9083 & 0.8959 & 0.8448 & 0.8656 & 0.8074 & 0.8048 \\
    TrafficFormer   & 0.4750 & 0.7366 & 0.4750 & 0.3320 & \textbf{0.9750} & \textbf{0.9789} & \textbf{0.9750} & \textbf{0.9746} & 0.8783 & 0.8801 & 0.8783 & 0.8785 & 0.8725 & 0.8487 & 0.8343 & 0.8288 \\
    \hline
    FlowletFormer   & \textbf{0.7050} & \textbf{0.7742} & \textbf{0.7050} & \textbf{0.6684} & 0.9650 & 0.9689 & 0.9650 & 0.9648 & \textbf{0.9183} & \textbf{0.9475} & \textbf{0.9183} & \textbf{0.9079} & \textbf{0.9109} & \textbf{0.8905} & \textbf{0.8866} & \textbf{0.8859} \\
    \bottomrule
    \end{tabular}
\caption{Comparison results on Browser, USTC-TFC, CIC-IDS2017, and CIC-IoT2022 datasets.}
\label{tab:metrics2}
\end{table*}

\section{Experiment}

\subsection{Experiment Setup}

\textbf{Pre-training Dataset.} In this work, approximately 30GB of unlabeled raw traffic data is used for pre-training. The dataset was sourced from three main repositories: ISCX-VPN2016 (NonVPN) \cite{DraperCharacterization}, CIC-IDS2017 (Monday) \cite{SharafaldinToward}, and the WIDE backbone dataset (January 1, 2024) \cite{ChoTraffic}. These datasets encompass a variety of network application scenarios and protocols, such as web browsing with HTTP, file downloads with FTP, email with SMTP, and video streaming with QUIC.

During pre-training dataset construction, we extract 64 consecutive bytes from the beginning of the Network Layer of each packet as the model input, in order to cover key information from the IP layer and above. Furthermore, \textbf{no randomization was applied to the pre-training dataset.} 
% Furthermore, the datasets provide backbone traffic with an extended time span, reflecting the overall trends in recent network behavior.

% \textbf{Fine-tuning Dataset.} We employ eight datasets for fine-tuning. These datasets correspond to seven different downstream tasks: \textbf{Service Type Identification} determines the specific service type associated with network traffic through traffic analysis, including ISCX-VPN (Service) \cite{DraperCharacterization} and ISCX-Tor2016 \cite{LashkariDMG17}. \textbf{Application Classification} identifies the application generating the traffic, including ISCX-VPN (App) \cite{DraperCharacterization}. \textbf{Website Fingerprinting} recognizes websites visited by users within encrypted traffic, including CSTNET-TLS \cite{Linetbert}. \textbf{Browser Classification} identifies the browser type used by analyzing traffic features, including Browser \footnote{The link address of the Browser dataset is https://drive.google.com/file/d/1wOdrfazbrcMDrL0NfA4GLoWegtPqkPj3.}. \textbf{Malware Classification} detects and categorizes traffic generated by malicious software, including USTC-TFC \cite{Wangmalware}. \textbf{Malicious Traffic Classification} identifies and differentiates various malicious network attack behaviors, including CIC-IDS2017 \cite{SharafaldinToward}. \textbf{IoT Classification} analyzes traffic characteristics to identify IoT device types or behavioral patterns, including CIC-IoT2022 \cite{DadkhahIoT}.

\textbf{Fine-tuning Dataset.} We employ 8 datasets for fine-tuning, corresponding to 7 different downstream tasks, including 
Service Type Identification (ISCX-VPN (Service) \cite{DraperCharacterization} and ISCX-Tor2016 \cite{LashkariDMG17}), 
Application Classification (ISCX-VPN (App) \cite{DraperCharacterization}),
{Website Fingerprinting (CSTNET-TLS \cite{Linetbert}),
Browser Classification (Browser \cite{LiuHXCL19FSNET}),
Malware Classification (USTC-TFC \cite{Wangmalware}), 
Malicious Traffic Classification (CIC-IDS2017 \cite{SharafaldinToward}),
and IoT Classification (CIC-IoT2022 \cite{DadkhahIoT}).

During fine-tuning dataset construction, we select the first five packets of each flow and extract 64 bytes starting from the Network Layer of each packet. To preserve privacy and mitigate potential biases, \textbf{we further anonymize the packets by applying IP Address\&Port randomization and TCP timestamp adjustments.} The pretraining and fine-tuning datasets are strictly separated to avoid data leakage.

\textbf{Evaluation Metrics.} We adopt classification accuracy (AC), precision (PR), recall (RC), and F1 score as the evaluation metrics for the experiments. More Experiment Setup details are shown in Appendix D.

\subsection{Comparison with State-of-the-Art Methods}
We compare FlowletFormer with various baselines and state-of-the-art methods. AppScanner \cite{TaylorSCM16}, BIND \cite{AlAdaptive}, and CUMUL \cite{PanchenkoWebsite} are based on ML models. DF \cite{SirinamIJW18DF}, FSNet \cite{LiuHXCL19FSNET}, GraphDapp \cite{ShenZZXD21GraphDapp} and Beauty \cite{SchusterST17Beauty} use DL models. ET-BERT \cite{Linetbert}, YaTC \cite{ZhaoYATC} and TrafficFormer \cite{ZhouTrafficFormer} are pre-training methods. All pretraining methods are trained on the same pre-training dataset and fine-tuning dataset. The ML and DL models are trained on fine-tuning dataset.

As shown in Table \ref{tab:metrics1} and \ref{tab:metrics2}, FlowletFormer outperforms all methods on 7 datasets. Especially in the Service Type Identification (ISCX-VPN Service, Tor) task, FlowletFormer attains an F1-score of 0.9364 and 0.9116, outperforming the second-best methods, FSNet and ET-BERT, by 3.1\% and 16.6\%, respectively. Even in the Malware Classification Task, FlowletFormer is only 1\% lower than the best performing method (TrafficFormer). In fact, this trivial difference largely varies due to random data splits; we show in Appendix E that FlowletFormer outperforms TrafficFormer in 3 of the other 5 splits using different random seeds on this dataset.
% The reason for this difference is dataset split. In additional evaluation, FlowletFormer outperformed TrafficFormer in 3 of the 5 random dataset splits (Appendix \ref{Appendix:Malware}).
% This demonstrates its ability to accurately identify specific service types in complex network environments. 
% These results validate FlowletFormer's capability to learn hierarchical flow representations that capture both intra-packet structural features and inter-packet behavioral semantics across diverse network environments.
These results demonstrate that FlowletFormer, as a pre-training model with a more specific design for traffic, can flexibly and effectively adapt to various traffic classification tasks in diverse network environments, indicating its promise in improving network management and security.

\subsection{Ablation Study}
To evaluate the contribution of different components in FlowletFormer, we conduct an ablation study on 2 datasets. Specifically, we systematically remove key components, including FlowLet, the Masked Field Model, the Flowlet Prediction Task, the Protocol Stack Alignment-Based Embedding Layer, and the Pre-training stage. The results are presented in Figure \ref{fig:ablation}. First, Flowlet significantly improves recall and overall F1 score by capturing fine-grained Flowlet dependencies, while the Masked Field Model helps the model learn structured contextual representations, leading to better generalization. The Flowlet Prediction Task contributes to flowlet-level continuity modeling, which is essential for robust classification. The Protocol Stack Alignment-Based Embedding Layer provides marginal improvements, indicating its role in refining hierarchical representations. Most notably, the Pre-training is the most important, as removing it leads to a drastic performance drop, highlighting its importance in learning general traffic representations. The details are provided in Appendix E.

\begin{figure}[h]
    \centering
    \subfloat[ISCX-VPN(APP)]
{\includegraphics[height=0.185\textwidth]{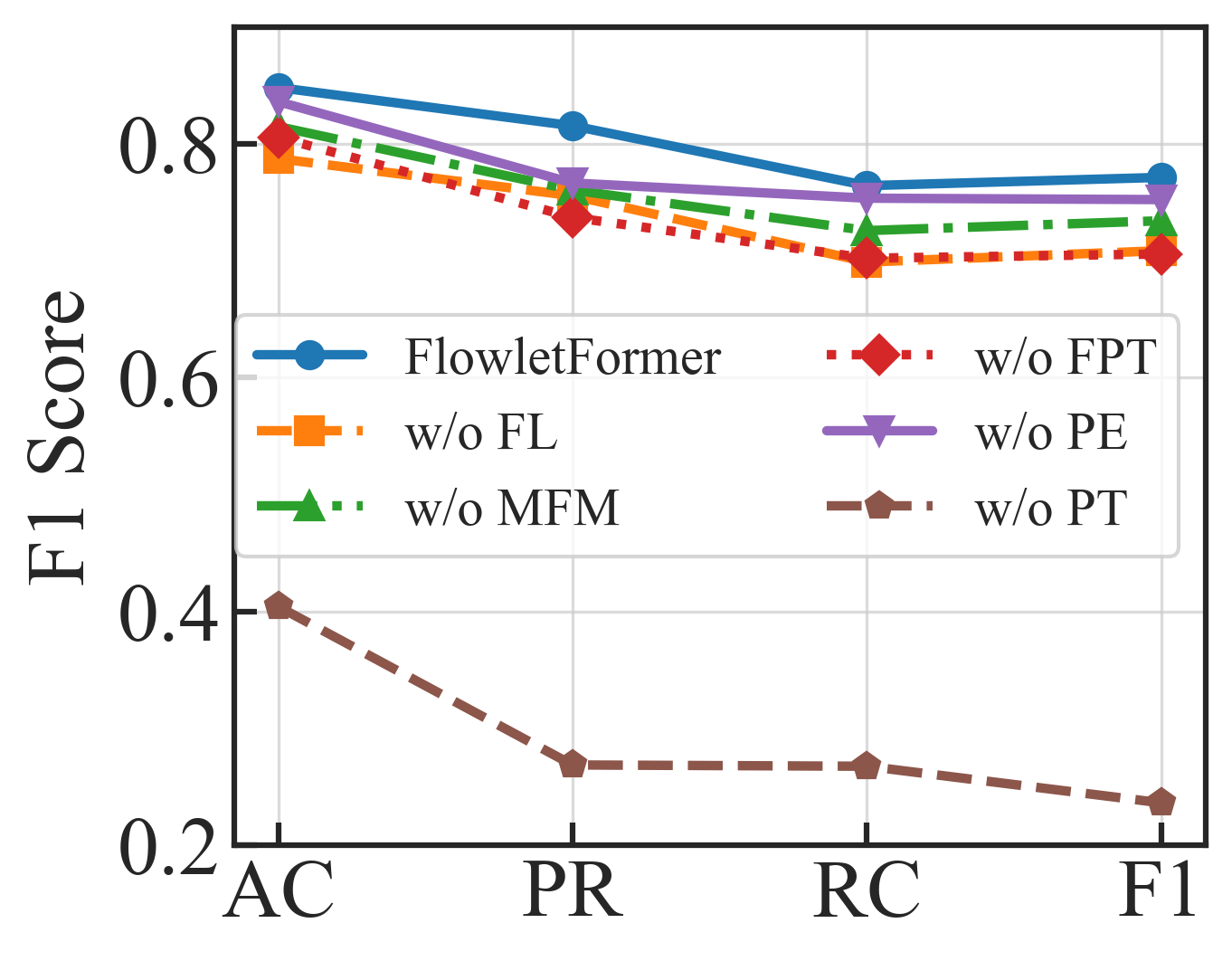} %
    \label{subfig:ablation1}
    }
    \subfloat[CSTNET-TLS]
    {\includegraphics[height=0.19\textwidth]{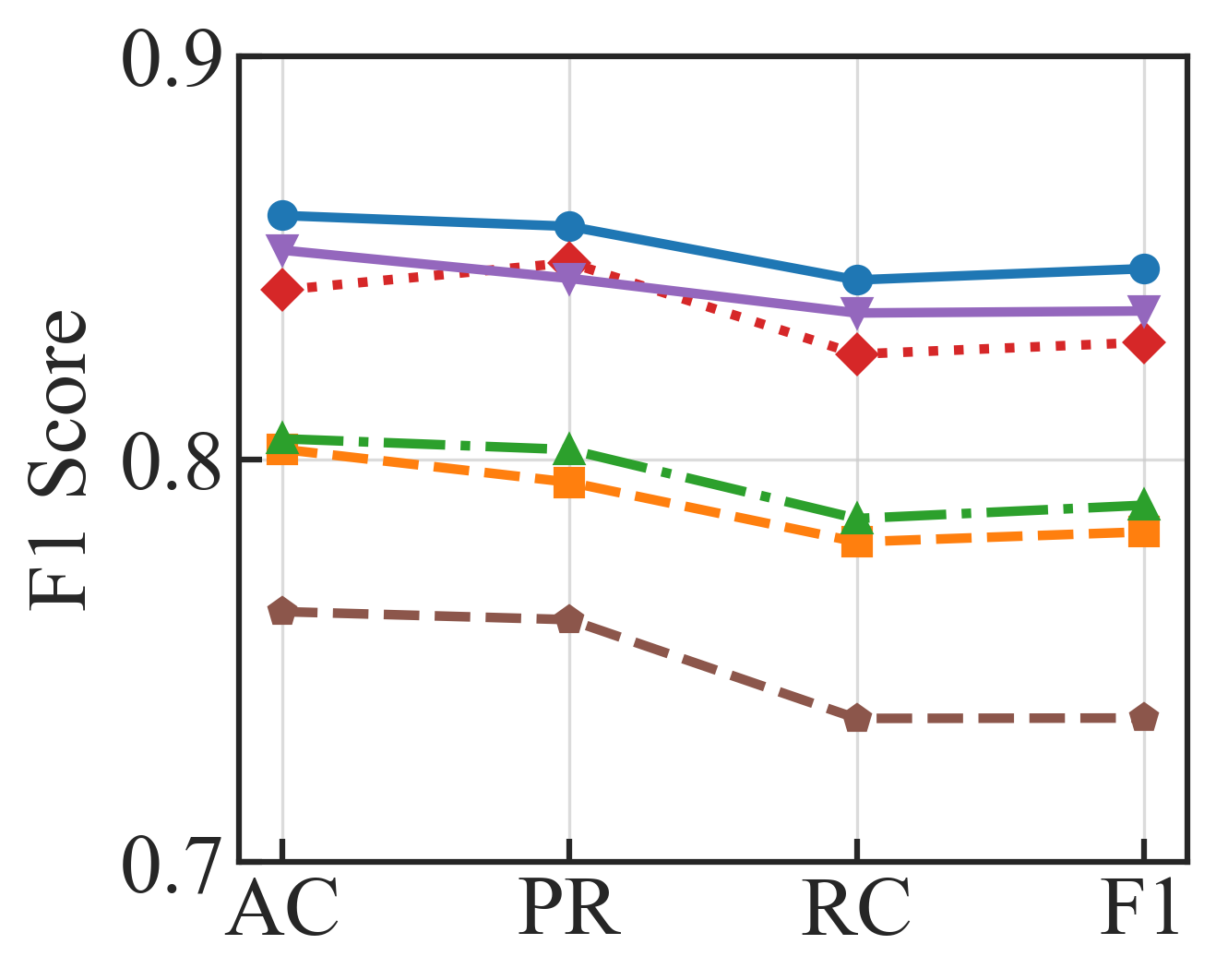}
    \label{subfig:ablation2}
    }
    \caption{\textbf{Ablation Study of key components in FlowletFormer.} The abbreviations are explained as follows: FL: Flowlet and Field Tokenization, MFM: Masked Field Model, FPT: Flowlet Prediction Task, PE: Protocol Stack Alignment-Based Embedding Layer, and PT: Pre-Training.}
    \label{fig:ablation}
\end{figure}

\begin{table*}[htbp]
    \small
    \centering
    \tabcolsep=0.1cm
    \begin{tabular}{c|ccc|ccc|ccc|ccc}
    \toprule
    Task & \multicolumn{3}{c|}{Flow Direction Inference} & \multicolumn{3}{c|}{Transport Protocols Recognition} & \multicolumn{3}{c|}{Sequence Awareness} & \multicolumn{3}{c}{Connection control Judgement}  \\
    \cline{1-13}
    Dataset & VPN & IDS & TFC & VPN & IDS & TFC & VPN & IDS & TFC & VPN & IDS & TFC \\
    \hline
    ET-BERT & 0.4366 & 0.7096 & 0.7412 & 0.9681 & 0.9767 & 0.9981 & 0.4165 & 0.6937 & 0.6203 & 0.9041 & 0.9975 & 0.9985  \\
    TrafficFormer & 0.0164 & 0.1059 & 0.1128 & 0.6753 & 0.9067 & 0.8912 & 0.3659 & 0.5261 & 0.3652 & 0.3904 & 0.9983 & 0.9978 \\
    \hline
    FlowletFormer   & \textbf{0.9313} & \textbf{0.9647} & \textbf{0.9196} & \textbf{1.0000} & \textbf{1.0000} & \textbf{1.0000} & \textbf{0.6987} & \textbf{0.7806} & \textbf{0.7579} & \textbf{0.9338} & \textbf{1.0000} & \textbf{1.0000} \\
    \bottomrule
    \end{tabular}
\caption{The performance comparison with other pre-training models on field understanding tasks.}
\label{tab:field}
\end{table*}

\subsection{Few-shot Analysis}
To further assess the effectiveness and robustness of FlowletFormer under few-shot conditions, we conduct experiments with varying data proportions on 2 datasets. Specifically, we use the full dataset as the reference and randomly sample 40\%, 20\%, and 10\% of the available data for few-shot training. Our few-shot evaluation on ISCX-VPN-App reveals FlowletFormer's superior data efficiency, maintaining F1 scores of 0.8009 (40\% data), 0.6224 (20\%), and 0.5813 (10\%) – surpassing state-of-the-art baselines by 12.7-46.1\% margins. Notably, while supervised methods (e.g., BIND/DF/FSNet) exhibit catastrophic performance under data scarcity (DF declines 46.1\% at 10\% data), our pre-training framework maintains performance through the traffic representation model, as evidenced in Figure \ref{fig:fewshot}. More results can be found in Appendix E.

\begin{figure}[h]
    \centering
    {\includegraphics[height=0.15\textwidth]{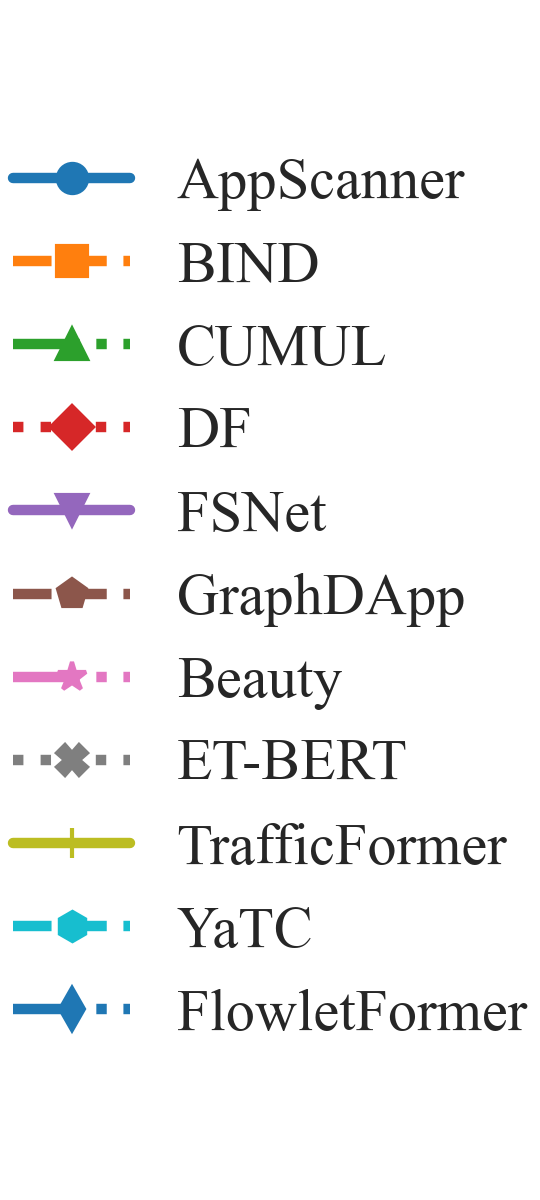}}
    \subfloat[ISCX-VPN(APP)]
    {\includegraphics[height=0.14\textwidth]{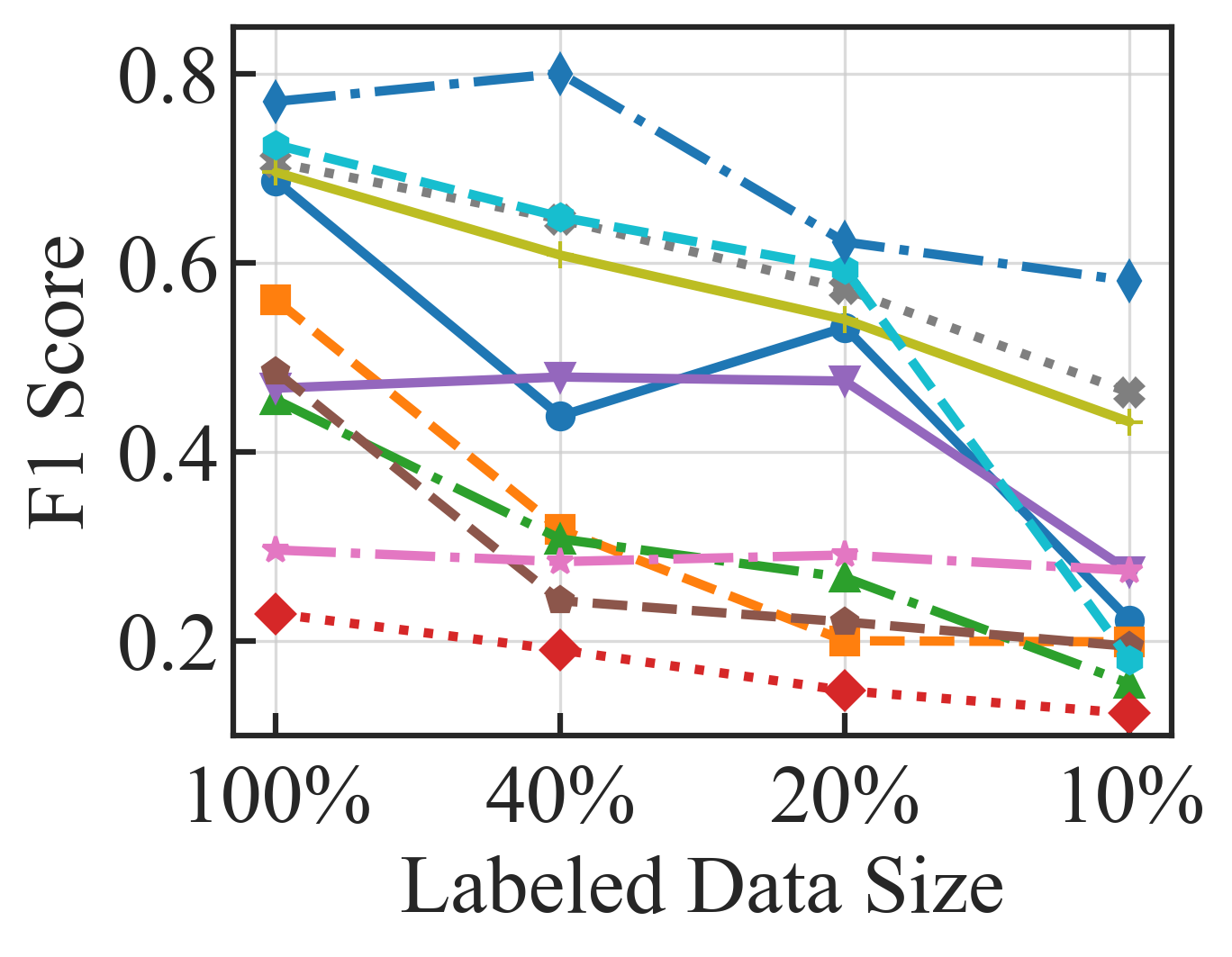} %
    \label{subfig:fewshot1}
    }
    \subfloat[CSTNET-TLS]
    {\includegraphics[height=0.14\textwidth]{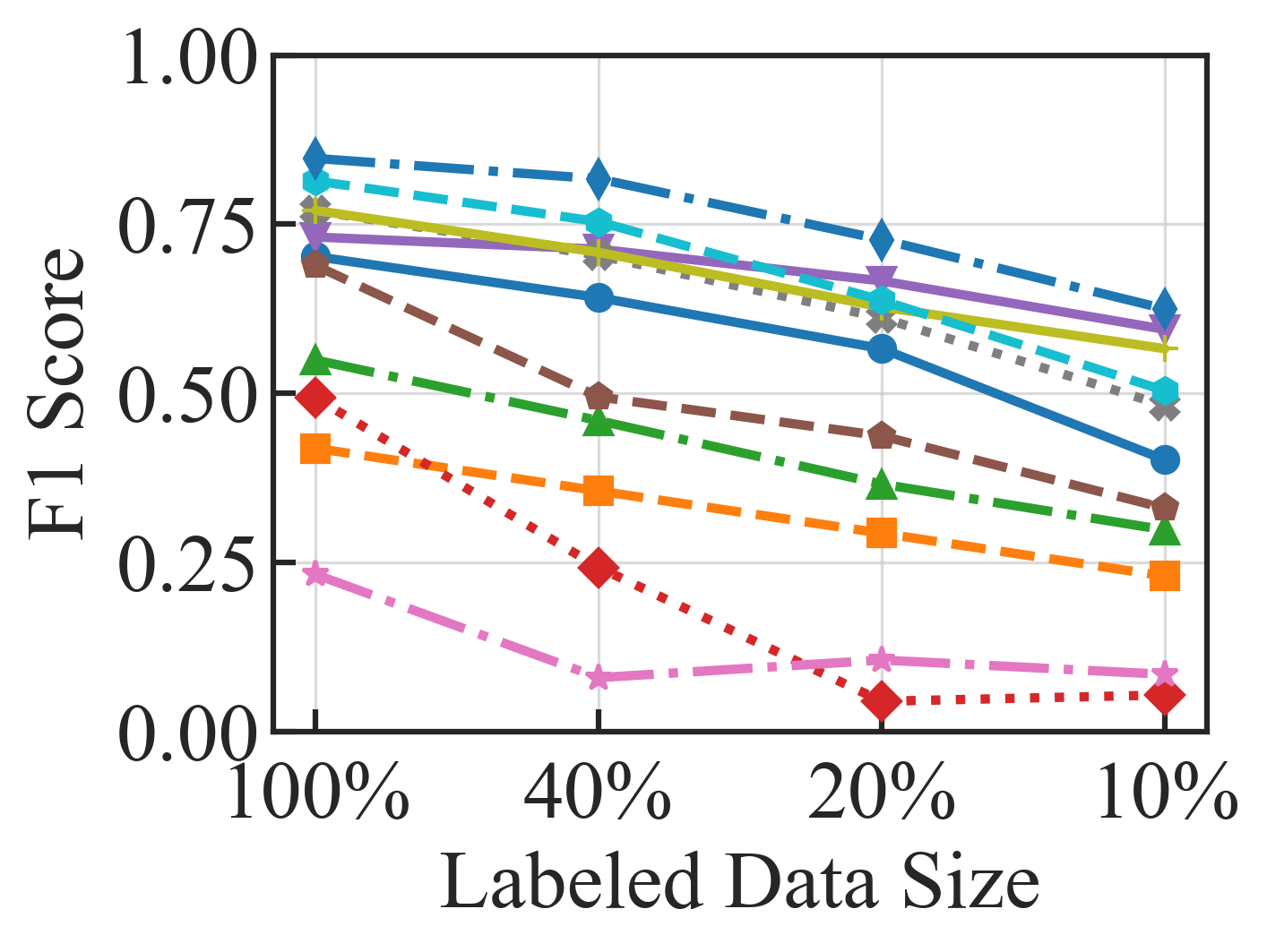}
    \label{subfig:fewshot2}
    }
    \caption{\textbf{Few-shot Analysis.}}
    \label{fig:fewshot}
\end{figure}

\subsection{Field Understanding Task}
We introduce multiple \textbf{Field Understanding Tasks} to assess whether the pre-trained model\textsuperscript{\ref{ft:pretrained}} comprehends general traffic patterns. These tasks require the model to predict key header fields within a packet in a given flow. Specifically, we evaluate the comprehension of the model in four tasks: the \textbf{Flow Direction Inference} task masks the source/destination IP and Ports, assessing the model’s ability to infer packet direction between entities based on contextual clues without direct address information; the \textbf{Transport Protocol Recognition} task focuses on masking the protocol field in the IP header, testing the model’s ability to identify the transport layer protocol (e.g., TCP, UDP, ICMP); the \textbf{Sequence Awareness} task masks the sequence number and acknowledgment number within the TCP header, challenging the model to infer packet order and flow continuity; the \textbf{Connection Control Judgment} task masks the flag fields in the TCP header, which denote the state of the connection, and evaluates the model’s ability to infer control signals like session establishment or termination. 

These tasks evaluate the model's ability to infer direction, protocol, sequence, and control, with performance measured in three datasets: ISCX VPN, CICIDS2017, and USTC-TFC. As shown in Table \ref{tab:field}, FlowletFormer outperforms two models in all tasks. The model’s ability to effectively infer Flow Direction, Transport Protocol, Sequence Awareness, and Connection Control across diverse datasets demonstrates its strong capacity for understanding the complex behavior of network traffic. 

\subsection{Word Analogies Similarity Analysis}
In NLP, word analogy tasks assess a model’s ability to capture semantic relationships between words. Through word analogy similarity analysis, we can validate whether a model has deeply understood the semantic relationships between words. Similarly, the port number analogy analysis can be used to evaluate the pre-trained model\textsuperscript{\ref{ft:pretrained}}, assessing its understanding of the functional and semantic relationships between network services. This capability reflects the model’s deep understanding of traffic patterns acquired during pretraining, without any downstream fine-tuning.

We apply cosine similarities between the embeddings of port numbers produced by the pre-trained model to examine the relationships among common HTTP–related ports (e.g., 80, 8080, 8000). Comparing ET-BERT with FlowletFormer (Table \ref{tab:Similarity}), we find that ET-BERT struggles to model port similarities, while FlowletFormer effectively captures these relationships, enhancing traffic classification performance. Appendix E provides more clarification.
\begin{table}[htbp]
    \small
    \centering
    \tabcolsep=0.1cm
    \begin{tabular}{c|cc|cc|cc}
    \toprule
    Port & \multicolumn{2}{c|}{80\&8080} & \multicolumn{2}{c|}{80\&8000} & \multicolumn{2}{c}{8080\&8000}  \\
    \hline
    Embedding & Word & Input & Word & Input & Word & Input \\
    \hline
    ET-BERT    & -0.0768 & 0.1094 & -0.0685 & 0.1331 & \textbf{0.0740} & 0.2438 \\
    FlowletFormer   & \textbf{0.0582} & \textbf{0.4019} & \textbf{0.0369} & \textbf{0.3993} & 0.0400 & \textbf{0.4289} \\
    \bottomrule
    \end{tabular}
\caption{Port Number Analogy Cosine Similarity about Word Embedding and Input Embedding.}
\label{tab:Similarity}
\end{table}

\subsection{Computational Cost and Complexity}
We analyze the time complexity of our method. Since our model is built upon the BERT-Base architecture, it shares the same pre-training time complexity as the two word-like methods, ET-BERT and TrafficFormer. Specifically, the complexity is:
\begin{equation}
\mathcal{O}(N \times B \times L \times (S^2 \cdot H + S \cdot H^2))
\end{equation}
where $N$ is the number of training steps, $B$ is the batch size, $L$ is the number of Transformer layers, $S$ is the input sequence length, and $H$ is the hidden size. 
We also measure the end-to-end runtimes of FlowletFormer during different phases of the train. Table~\ref{tab:computation_flowletformer} summarizes these results.

\begin{table}[htbp]
\centering
\small
\tabcolsep=0.06cm
\begin{tabular}{lccccc}
\toprule
\textbf{Phase} & \textbf{GPUs} & \textbf{Time} & \textbf{Unit/Granularity} & \textbf{GPU Memory (GB)} \\
\midrule
Pre-training  & 6 & 42 h  & 75.67 s / 100 steps & 28 \\
Fine-tuning   & 1 & 1,153 s & 57.65 s / epoch    & 17 \\
Inference     & 1 & –     & 150.04 samples/sec  & –  \\
\bottomrule
\end{tabular}
\caption{Computational Efficiency Across Different Phases about FlowletFormer. More detail are shown in Appendix E.}
\label{tab:computation_flowletformer}
\end{table}

\section{Conclusion}
In this paper, we propose FlowletFormer, a BERT-based pre-training model designed for network traffic analysis. By introducing a Coherent Behavior-Aware Traffic Representation Model, a Protocol Stack Alignment-Based Embedding Layer, and Field-Specific and Context-Aware Pretraining Tasks, FlowletFormer effectively captures behavioral patterns, hierarchical protocol semantics, and inter-packet contextual relationships among traffic data. The experimental results demonstrate its superiority over existing methods in traffic classification. 

FlowletFormer improves network traffic classification, but challenges remain. Future work includes adapting to evolving traffic patterns, enhancing robustness against adversarial attacks, incorporating multi-modal data, and optimizing computational efficiency for real-time deployment. Addressing these issues will strengthen its role in network security and traffic classification.

\bibliography{aaai2026}


% \input{ReproducibilityChecklist}
\clearpage

\input{appendix}

\end{document}

%% file: appendix.tex
\section{Appendix}
\subsection{Preliminary Analysis}
\label{Appendix:preliminary}
We conduct three in-depth analyses to examine the limitations of existing tokenization and segmentation strategies for traffic data representation:

\textbf{First}, we investigate the applicability of sub-word tokenization techniques, which are widely adopted in NLP to reduce vocabulary size. Methods such as ET-BERT and TrafficFormer directly employ byte-pair encoding (BPE)  to tokenize traffic data. To evaluate its effectiveness in this domain, we apply sub-word tokenization to the CICIDS2017-Monday dataset. Statistical results show that \textbf{less than 1\% of the tokens are segmented into subwords} (e.g., \colorbox{lightgray}{\texttt{1df8}} into \colorbox{lightgray}{\texttt{1df}} and \colorbox{lightgray}{\texttt{\#8}}), indicating that most tokens remain unsplit. This suggests that such techniques may offer limited granularity and expressiveness when applied to traffic data.

\textbf{Second}, we identify a critical limitation in how existing methods often overlook the hierarchical structure of packet data. Unlike natural language, network packets are composed of protocol layers with distinct semantics (e.g., IP, TCP, application layer). Flattening these layered structures into a linear token sequence may result in a loss of structural information, hindering the model’s capacity to learn meaningful representations.

\textbf{Third}, we analyze burst segmentation in the CICIDS2017-Monday dataset. Bursts are commonly used to reflect traffic behaviors in application layer. We compute the cumulative distribution function (CDF) of the number of packets per burst and find that \textbf{approximately 65\% of bursts consist of only a single packet}. This suggests that the dataset contains a high proportion of extremely short bursts, which limits the temporal context available for modeling. Moreover, since existing pretraining tasks (e.g., Same-origin Burst Prediction) often rely on splitting bursts into sub-bursts or sub-packets, the semantic connections between these sub-units are likely to be weak. Consequently, the model may struggle to learn robust traffic patterns from such limited contexts. 

\subsection{More Details of Our Method}
\subsubsection{Flow Construction}
\label{Appendix:flow}
To construct semantically meaningful flows from raw packet data, we apply protocol-specific rules according to standard practices outlined in RFCs and previous works. The flow construction process is based on the five-tuple: \texttt{{srcIP, dstIP, srcPort, dstPort, protocol}}, with additional considerations depending on the transport layer protocol.

We apply protocol-specific rules based on both packet semantics and timeout heuristics. As shown in Table~\ref{tab:flowconstructionrules}, different protocols adopt distinct termination and reinitialization criteria. For instance, TCP flows are explicitly closed by a four-way handshake or reset flag, while UDP and ICMP rely on timeout-based or field-change-based segmentation. These rules help segment raw traffic into coherent flow units for downstream analysis.

\begin{table*}[h]
\caption{\textbf{Protocol-specific Rules for Flow Construction.}}
\vskip 0.15in
\scriptsize
\centering
\renewcommand\arraystretch{1.1}
\begin{tabular}{c|l|l}
\toprule
\textbf{Protocol} & \textbf{Flow Termination Condition} & \textbf{New Flow Trigger} \\
\midrule
TCP & 
\begin{tabular}[c]{@{}l@{}}
Four-way Handshake (FIN + FIN + ACK) \\
Connection Reset (RST packet) \\
Active Timeout (Flow duration exceeds 1800s)
\end{tabular} &
\begin{tabular}[c]{@{}l@{}}
New SYN + ACK Connection \\
Active Timeout Expiration
\end{tabular} \\
\midrule
UDP & 
\begin{tabular}[t]{@{}l@{}}
Inactive Timeout (Flow duration exceeds 15s)
\end{tabular} &
Inactive Timeout Expiration \\
\midrule
\multirow{2}{*}{ICMP} & 
\begin{tabular}[t]{@{}l@{}}
Change in ICMP \texttt{Type} \\
Change in ICMP \texttt{Code}
\end{tabular} &
\multirow{2}{*}{Any change in \texttt{Type} or \texttt{Code}} \\
\midrule
Others & 
\begin{tabular}[t]{@{}l@{}}
Flow duration exceeds 1800 seconds
\end{tabular} &
Timeout Expiration \\
\bottomrule
\end{tabular}
\label{tab:flowconstructionrules}
\end{table*}

\subsubsection{Flowlet Generation}
\label{Appendix:flowlet}
After flow construction, we perform the Flowlet Generation. We also describe it in Algorithm \ref{alg:flowletsplit} The Flowlet Generation Algorithm dynamically partitions a flow into flowlets based on inter-packet arrival time. It operates as follows:

\begin{itemize}
    \item \textbf{Initialization}: For each network flow $\boldsymbol{F} = \{\text{pkt}_1, \dots, \text{pkt}_n\}$ with timestamps $\{\tau_1, \dots, \tau_n\}$, we compute the average inter-arrival time of the first three packets, i.e., $\theta_3 = \frac{1}{2}[(\tau_2 - \tau_1) + (\tau_3 - \tau_2)]$. This value is used as the initial threshold $\theta$ for segmentation. If $n \leq 3$, the entire flow is treated as a single Flowlet.
    \item \textbf{Segmentation}: For each subsequent packet $\text{pkt}_i$ ($i > 3$), we calculate the inter-arrival time $t_i = \tau_i - \tau_{i-1}$. If $t_i > \theta_{i-1}$, we create a segmentation: the previous packet $\text{pkt}_{i-1}$ ends the current Flowlet $\mathcal{F}_j$, and $\text{pkt}_i$ begins a new one $\mathcal{F}_{j+1}$. Otherwise, $\text{pkt}_i$ is appended to the current $\mathcal{F}_j$.
    \item \textbf{Threshold Update}: After each decision, we update the threshold $\theta_i$ using all observed inter-arrival times up to index $i$, i.e., $\theta_i = \frac{1}{|W_i|} \sum_{t \in W_i} t$, where $W_i$ is the window of past IATs. This allows the threshold to adapt dynamically to local flow patterns.
\end{itemize}

This adaptive thresholding approach allows the segmentation process to adjust to diverse traffic dynamics. For instance, traffic patterns such as HTTP request-response cycles or video streaming often exhibit short bursts followed by longer silent gaps. By capturing such timing structures, Flowlet segmentation enables the model to better align with the logical behavior units within network communication, thus enhancing the semantic granularity of traffic representation.

\begin{algorithm}[h]
   \caption{Flowlet Generation}
   \label{alg:flowletsplit}
\begin{algorithmic}[1]
   \STATE \textbf{Input:} Flow $\boldsymbol{F} = \{\text{pkt}_1, \dots, \text{pkt}_n\}$ with arrival timestamps $\{\tau_1, \dots, \tau_n\}$
   \STATE \textbf{Output:} Flowlets $\{\mathcal{F}_1, \dots, \mathcal{F}_k\}$
   \STATE \textbf{Initialize:} $\mathcal{F} \gets \{\text{pkt}_1\}$, $W \gets \emptyset$, $\boldsymbol{flowlets} \gets \emptyset$
   \FOR{$i \gets 2$ \textbf{to} $n$}
       \STATE $t_i \gets \tau_i - \tau_{i-1}$
       \IF{$i > 3$ \textbf{and} $t_i > \theta_{i-1}$}
           \STATE Append $\mathcal{F}$ to $\boldsymbol{flowlets}$
           \STATE $\mathcal{F} \gets \{\text{pkt}_i\}$
       \ELSE
           \STATE Append $\text{pkt}_i$ to $\mathcal{F}$
       \ENDIF
       \STATE Append $t_i$ to $W$
       \STATE $\theta_i \gets \frac{1}{|W|} \sum_{t \in W} t$
   \ENDFOR
   \STATE Append remaining $\mathcal{F}$ to $\boldsymbol{flowlets}$
\end{algorithmic}
\end{algorithm}

\subsubsection{Key Protocol Header Fields in Masked Field Model}
\label{Appendix:fields}
Table~\ref{keyfields} lists the key fields commonly found in standard network protocols. These fields carry rich semantic and structural information that can be leveraged by traffic analysis models.

For example, fields such as IP addresses, port numbers, and protocol types provide fundamental information about the directionality and service type of a packet, helping models distinguish between client-server roles or application types. 

Sequence Number and Acknowledgment Number in the TCP header reflect the transmission order and reliability mechanisms of the protocol, offering temporal cues to infer packet sequences and session continuity. 

The Total Length field, which indicates the size of an entire packet, has been demonstrated to serve as an effective signature for encrypted traffic classification in prior studies.

Furthermore, TCP control flags (e.g., SYN, ACK, FIN, RST) encode connection state transitions (e.g., handshake, termination), enabling models to learn flow dynamics and session boundaries. 

Similarly, ICMP's Type and Code fields identify message semantics (e.g., echo request/reply, destination unreachable), while the minimal set of fields in UDP (primarily source and destination ports) still conveys important endpoint semantics.

\begin{table}[h]
\caption{Key fields in common protocol.}
\label{keyfields}
\begin{center}
\begin{scriptsize}
\begin{tabular}{c|l}
\toprule
\textbf{Protocol} & \textbf{Key Fields} \\
\midrule
IP      & Version, Total Length, Protocol, IP Address  \\
\midrule
\multirow{2}{*}{TCP}     & Port Number, Sequence Number, Flag \\
        & Acknowledgment Number, Window Size \\
\midrule
UDP     & Port Number \\
\midrule
ICMP    & Type, Code  \\
\bottomrule
\end{tabular}
\end{scriptsize}
\end{center}
\end{table}

% \subsection{Implementation of Baselines}
% \label{Appendix:baseline}

\subsection{More Details in Experiment Setup}
\label{Appendix:experiment}
\subsubsection{More Details in Pre-training Dataset Construction}
\label{Appendix:Pre-trainingDataset}

We describe the data preprocessing pipeline used during the pre-training stage of FlowletFormer.

\textbf{Flow Construction.}
We first parsed raw PCAP files to construct flows based on five-tuples and protocol-specific rules which ensure semantically coherent flow boundaries. Each flow was saved as an individual PCAP file for subsequent processing.

\textbf{Flowlet Segmentation.}
To better reflect the temporal structure and traffic behavior from application layer, we further segmented each flow into multiple flowlets. Specifically, we calculated inter-packet arrival times (IATs) and initiated a new flowlet whenever the IAT exceeded a threshold. This segmentation captures distinct behavioral units within each flow and enables the model to learn fine-grained communication patterns.

\textbf{Tokenization.}
For each packet in a flowlet, we removed the Ethernet header and retained the first 64 bytes starting from the network layer. These bytes were tokenized using Field Tokenization, where individual fields in protocol headers (e.g., IP version, TTL, TCP flags) are identified and converted into semantically meaningful tokens. This tokenization approach preserves protocol semantics while producing a consistent and structured input format for the model.

Table~\ref{tab:pre-trainingdataset} summarizes the pre-training datasets used in this work, including their sizes, number of flows, and supported protocols.

\begin{table*}[h]
\caption{Overview of Pre-training Datasets.}
\vskip 0.15in
    \scriptsize
    \centering
    \tabcolsep=0.1cm
    \renewcommand\arraystretch{0.8}
    \begin{tabular}{c|c|c|l}
    \toprule
    \textbf{Dataset} & \textbf{Size} & \textbf{Flow Number} & \textbf{Protocol} \\
    \midrule
    ISCX-VPN2016-NonVPN & 10.4G & 74,184 & TLS1.2, SFTP, SSDP, SNMP, NTP, MDNS, HTTP, GQUIC... \\
    CIC-IDS2017-Monday & 11G & 303,436 & HTTP, HTTPS, FTP, SSH, email protocols... \\
    WIDE-2024/1/1 & 9.6G & 2,322,172 & FTP, SSH, IPSec, HTTP, TLS1.2, TLS1.3, GRE, Email Protocol... \\
    \bottomrule
    \end{tabular}
    \label{tab:pre-trainingdataset}
\vskip -0.1in
\end{table*}
\subsubsection{More Details in Fine-tuning  Dataset Construction}
\label{Appendix:Fine-tuningDataset}
To ensure fair comparison and reproducibility, we describe the data preprocessing pipeline used during the fine-tuning stage of FlowletFormer.

\textbf{Data Collection and Filtering.}  
We collected raw PCAP files corresponding to the eight downstream tasks. Flows were constructed based on five-tuples $(\text{srcIP}, \text{dstIP}, \text{srcPort}, \text{dstPort}, \text{protocol})$, and each flow was saved as a separate PCAP file. 

Flows were then organized by traffic category. To facilitate manageable storage and training, large files were split into smaller ones (approximately 1,000 packets each). Categories with fewer than 10 samples were discarded, and a maximum of 500 samples per class was retained to ensure balanced representation.

\textbf{Data Anonymization and Randomization.}  
To mitigate the risk of shortcut learning and reduce the model’s dependence on protocol-specific artifacts, we performed the following anonymization steps on each flow:

\begin{itemize}
    \item Replaced all IP addresses with randomly generated addresses;
    \item Randomized source and destination ports while preserving client/server roles;
    \item Adjusted TCP timestamps by introducing a random base time, but preserving the relative inter-packet timing.
\end{itemize}

\textbf{Tokenization.}  
We selected the first five packets of each flow and converted their contents to input tokens. Each packet was tokenized by retaining the first 64 tokens.

Table~\ref{tab:finetuningdataset} provides an overview of all downstream tasks used for fine-tuning FlowletFormer, including dataset names, number of flows, number of classes, and example labels.

\begin{table*}[h]
\caption{Overview of Fine-Tuning Tasks and Datasets.}
\vskip 0.15in
    \scriptsize
    \centering
    \tabcolsep=0.1cm
    \renewcommand\arraystretch{0.9}
    \begin{tabular}{c|c|c|c|l}
    \toprule
    \textbf{Task} & \textbf{Dataset} & \textbf{Flow Number} & \textbf{Class Number} & \textbf{Label} \\
    \midrule
    \multirow{2}{*}{Service Type Identification}  & ISCX-VPN (Service) & 1,500 & 6 & VPN-Chat,VPN-Email,VPN-Ftp... \\
    & ISCX-Tor2016 & 2,922 & 8 & Audio, Browsing, Chat... \\
    \midrule
    Application Classification & ISCX-VPN (App) & 3,289 & 10 & VPN-Youtube,VPN-Voipbuster,VPN-Vimeo... \\
    \midrule
    Website Fingerprinting & CSTNET-TLS & 46,375 & 120 & acm.org,adobe.com,alibaba.com... \\
    \midrule
    Browser Classification & Browser & 2,000 & 4 & Chrome,Firefox,Internet,UC \\
    \midrule
    Malware Classification & USTC-TFC & 8,000 & 16 & Miuref,FTP,Gmail... \\
    \midrule
    Traffic Classification & CIC-IDS2017 & 6,000 & 12 & Benign,Botnet,DDoS... \\
    \midrule
    IoT Classification & CIC-IoT2022 & 4,931 & 12 & Attack\_Flood,Idle,Interaction\_Audio... \\
    \bottomrule
    \end{tabular}
    \label{tab:finetuningdataset}
\vskip -0.1in
\end{table*}

\subsubsection{More Details in Implementation}
\label{Appendix:implementation}
In this experiment, we employ multi-GPU parallel in pre-training. A total of six GPUs are used for distributed training, with a batch size set to 16, resulting in an overall batch size of 96. The total number of training steps is 200,000, with model checkpoints saved every 10,000 steps. The Adam optimizer is chosen, with an initial learning rate of 2e-5 and a warm-up ratio of 0.1 to ensure stability during the initial stages of training. 

To maintain consistency with pre-training, the fine-tuning data is processed in the same input format as the pre-training data. The packets in the flowlets are directly concatenated without [SEP] token for separation, meaning all tokens share the same segment identifiers. During the fine-tuning stage, we select the first five packets of each network flow as the model input and extract the first 64 tokens following the Ethernet header of each packet. The dataset is split into train/validation/test sets with an 8:1:1 ratio. The model was trained for up to 20 epochs on each dataset using the AdamW optimizer with a learning rate of 6e-5, with early stopping triggered if the F1 score did not improve for 4 consecutive epochs.

The proposed method is implemented using PyTorch 2.3.1 and UER and trained on a server with 8 NVIDIA Tesla V100S GPUs.

To comprehensively evaluate the performance of classification models, we adopt widely used metrics, accuracy (AC), precision (PR), recall (RC), and F1 score (F1).

In our evaluation, precision, recall, and F1 score are macro-averaged to ensure equal consideration of all classes regardless of their frequency.

\subsection{Additional Evaluation: Malware Dataset}
\label{Appendix:Malware}
We further investigated the 1\% lower F1 score of FlowletFormer versus TrafficFormer on the Malware classification task. Minor variations (like 1\% difference) on this dataset can occasionally arise from factors like specific data splits. To this end, we generated 5 random dataset splits by varying the random seed, and found that in 3 cases, FlowletFormer either outperformed or matched TrafficFormer on the Malware dataset. Table~\ref{tab:malware-seeds} reports the F1 scores under each seed.

\begin{table}[h]
  \caption{F1 scores on the Malware dataset under five random splits.}
  \label{tab:malware-seeds}
  \centering
  \scriptsize
  \begin{tabular}{c|cc}
    \toprule
    \textbf{Seed} & \textbf{TrafficFormer} & \textbf{FlowletFormer} \\
    \midrule
    SEED1 & 0.9753 & \textbf{0.9962} \\
    SEED2 & \textbf{0.9900} & 0.9615 \\
    SEED3 & 0.9764 & \textbf{0.9789} \\
    SEED4 & 0.9740 & \textbf{0.9776} \\
    SEED5 & \textbf{0.9788} & 0.9776 \\
    \bottomrule
  \end{tabular}
\end{table}

\subsection{More Ablation Study}
\label{Appendix:ablation}
To support the figures in the main text and further illustrate the robustness of our approach, we provide complete numerical results of the ablation study across all eight downstream datasets, as shown in Table~\ref{tab:ablation1} and Table~\ref{tab:ablation2}.

To thoroughly investigate the contribution of each component in \textbf{FlowletFormer}, we conducted a series of ablation experiments. The results in Table~\ref{tab:ablation1} and Table~\ref{tab:ablation2} report the performance of the full model and various degraded versions, where specific modules were removed.

\vspace{0.5em}
\textbf{Impact of Flowlet and Field Tokenization (FL).}  
Removing the Flowlet and Field Tokenization module (\texttt{w/o FL}) led to significant performance drops on most datasets. In this variant, the traffic representation and tokenization revert to the burst and BPE tokenization, which is consistent with the approach used in ET-BERT. For example, on the ISCX-Tor2016 dataset, the accuracy decreased from 0.9215 to 0.8328 and the F1-score from 0.9116 to 0.6924. The effect is even more pronounced on the Browser dataset, where accuracy dropped from 0.7050 to 0.3700 and F1-score from 0.6684 to 0.3099. These results highlight the critical role of Flowlet segmentation and field-aware tokenization in capturing temporal dependencies and contextual coherence within sessions. By introducing Flowlets, the model learns to represent traffic in a behavior-aware manner, which facilitates more robust classification of dynamic network flows.

\vspace{0.5em}
\textbf{Impact of Masked Field Model (MFM).}  
The removal of the masked field modeling task (\texttt{w/o MFM}) has dataset-specific effects. For instance, on the ISCX-VPN(Service) dataset, accuracy dropped dramatically from 0.9400 to 0.5467, indicating that MFM plays a critical role in modeling datasets with rich and structured protocol field information. It likely helps the model capture inter-field dependencies and learn which fields are important for traffic differentiation. In contrast, datasets like CSTNET-TLS and CIC-IDS2017 showed less degradation, suggesting that those tasks are less sensitive to fine-grained field semantics.

\vspace{0.5em}
\textbf{Impact of Flowlet Prediction Task (FPT).}  
Removing the Flowlet Prediction Task (\texttt{w/o FPT}) caused performance degradation across several datasets, though less severe than \texttt{w/o FL} or \texttt{w/o MFM}. For example, in ISCX-Tor2016, accuracy dropped from 0.9215 to 0.9044 and F1-score from 0.9116 to 0.8429. This indicates that FPT serves as an effective auxiliary task, guiding the model to learn patterns in the temporal evolution of traffic flows, which indirectly enhances downstream classification.

\vspace{0.5em}
\textbf{Impact of Protocol Stack Alignment-Based Embedding (PE).}  
The removal of the protocol embedding layer (\texttt{w/o PE}) resulted in a consistent but relatively moderate drop across datasets. For instance, F1-scores dropped by 2–4\%. This suggests that while PE enhances the model’s ability to capture protocol-layer semantics, it is not the main performance bottleneck.

\vspace{0.5em}
\textbf{Impact of Pretraining (PT).}  
Eliminating the pretraining stage (\texttt{w/o PT}) caused catastrophic performance degradation on all datasets. For example, on ISCX-VPN(Service), accuracy fell from 0.9400 to 0.5467 and F1-score from 0.9364 to 0.3949. These results emphasize the essential role of pretraining in learning generalizable traffic representations and initializing the model with better parameter priors for downstream tasks.

\begin{table*}[ht]
\centering
\caption{\textbf{Ablation study results on ISCXVPN2016, ISCX-Tor2016, and CSTNET-TLS 1.3 datasets.} The abbreviations are explained as follows, FL: Flowlet and Field Tokenization, MFM: Masked Field Model, FPT: Flowlet Prediction Task, PE: Protocol Stack Alignment-Based Embedding Layer and PT: Pre-Training.}
\scriptsize
\renewcommand\arraystretch{1.1}
\setlength{\tabcolsep}{1.5pt}
\begin{tabular}{l|cccc|cccc|cccc|cccc}
\toprule
\textbf{Dataset} & \multicolumn{4}{c|}{ISCX-VPN(Service)} & \multicolumn{4}{c|}{ISCX-Tor2016} & \multicolumn{4}{c|}{ISCX-VPN(App)} & \multicolumn{4}{c}{CSTNET-TLS} \\
\midrule
\textbf{Metric}  & AC & PR & RC & F1 & AC & PR & RC & F1 & AC & PR & RC & F1 & AC & PR & RC & F1 \\
\midrule
FlowletFormer & \textbf{0.9400} & \textbf{0.9471} & \textbf{0.9277} & \textbf{0.9364} & \textbf{0.9215} & \textbf{0.9263} & \textbf{0.9043} & \textbf{0.9116} & \textbf{0.8480} & \textbf{0.8153} & \textbf{0.7641} & \textbf{0.7712} & \textbf{0.8605} & \textbf{0.8578} & \textbf{0.8445} & \textbf{0.8473} \\
w/o FL & 0.9133 & 0.9077 & 0.8983 & 0.8995 & 0.8328 & 0.6978 & 0.6892 & 0.6924 & 0.7872 & 0.7555 & 0.6988 & 0.7085 & 0.8025 & 0.7943 & 0.7795 & 0.7820 \\
w/o MFM & 0.5467 & 0.5429 & 0.5323 & 0.4830 & 0.4505 & 0.1790 & 0.3300 & 0.2304 & 0.8146 & 0.7604 & 0.7257 & 0.7341 & 0.8051 & 0.8024 & 0.7853 & 0.7886 \\
w/o FPT & 0.9133 & 0.8936 & 0.9138 & 0.9010 & 0.9044 & 0.8189 & 0.9114 & 0.8429 & 0.8055 & 0.7370 & 0.7021 & 0.7057 & 0.8499 & 0.8597 & 0.8327 & 0.8372 \\
w/o PE & 0.9000 & 0.9087 & 0.8656 & 0.8804 & 0.9044 & 0.8428 & 0.9098 & 0.8653 & 0.8359 & 0.7664 & 0.7534 & 0.7522 & 0.8519 & 0.8449 & 0.8363 & 0.8368 \\
w/o PT & 0.5467 & 0.4278 & 0.4278 & 0.3949 & 0.1706 & 0.0213 & 0.1250 & 0.0364 & 0.4043 & 0.2689 & 0.2678 & 0.2365 & 0.7622 & 0.7602 & 0.7357 & 0.7358 \\
\bottomrule
\end{tabular}
\label{tab:ablation1}
\end{table*}

\begin{table*}[ht]
\centering
\caption{Ablation study results on Browser, USTC-TFC, CIC-IDS2017, and CIC-IoT2022 datasets.}
\scriptsize
\renewcommand\arraystretch{1.1}
\setlength{\tabcolsep}{1.5pt}
\begin{tabular}{l|cccc|cccc|cccc|cccc}
\toprule
\textbf{Dataset} & \multicolumn{4}{c|}{Browser} & \multicolumn{4}{c|}{USTC-TFC} & \multicolumn{4}{c|}{CIC-IDS2017} & \multicolumn{4}{c}{CIC-IoT2022} \\
\midrule
\textbf{Metric} & AC & PR & RC & F1 & AC & PR & RC & F1 & AC & PR & RC & F1 & AC & PR & RC & F1 \\
\midrule
FlowletFormer & \textbf{0.7050} & 0.7742 & \textbf{0.7050} & \textbf{0.6684} & \textbf{0.9675} & \textbf{0.9713} & \textbf{0.9675} & \textbf{0.9675} & \textbf{0.9183} & \textbf{0.9475} & \textbf{0.9183} & \textbf{0.9079} & \textbf{0.9109} & \textbf{0.8905} & \textbf{0.8866} & \textbf{0.8859} \\
w/o FL & 0.3700 & 0.2787 & 0.3700 & 0.3099 & 0.9600 & 0.9680 & 0.9600 & 0.9598 & 0.8850 & 0.8870 & 0.8850 & 0.8835 & 0.8401 & 0.7881 & 0.7936 & 0.7875 \\
w/o MFM & 0.6600 & 0.6006 & 0.6600 & 0.5976 & 0.9650 & 0.9723 & 0.9650 & 0.9653 & 0.4505 & 0.1790 & 0.3300 & 0.2304 & 0.8968 & 0.8506 & 0.8543 & 0.8473 \\
w/o FPT & 0.6850 & \textbf{0.7932} & 0.6850 & 0.6428 & 0.9663 & 0.9696 & 0.9663 & 0.9658 & 0.9044 & 0.8189 & 0.9114 & 0.8429 & 0.9049 & 0.8765 & 0.8788 & 0.8736 \\
w/o PE & 0.6800 & 0.7486 & 0.6800 & 0.6745 & 0.9650 & 0.9689 & 0.9650 & 0.9648 & 0.9044 & 0.8428 & 0.9098 & 0.8653 & 0.8988 & 0.8660 & 0.8593 & 0.8587 \\
w/o PT & 0.2700 & 0.3138 & 0.2700 & 0.1387 & 0.9563 & 0.9680 & 0.9562 & 0.9571 & 0.1706 & 0.0213 & 0.1250 & 0.0364 & 0.8664 & 0.8073 & 0.8174 & 0.8089 \\
\bottomrule
\end{tabular}
\label{tab:ablation2}
\end{table*}

\subsection{More Few-shot Analysis}
\label{Appendix:fewshot}
To evaluate the capability of FlowletFormer under data-scarce conditions, we conduct a few-shot learning analysis. The results are reported in Table~\ref{tab:fewshot1} and Table~\ref{tab:fewshot2}. As shown, FlowletFormer achieves competitive performance under full supervision (100\% training data). More importantly, it consistently maintains relatively high F1-scores even when the amount of training data is significantly reduced.

For example, on the ISCX-VPN(Service) dataset, FlowletFormer achieves an F1-score of 0.8106 using only 10\% of the training data, significantly outperforming traditional models such as AppScanner and BIND. This indicates the strong generalization ability of FlowletFormer in few-shot settings.

However, on the Browser dataset, the performance of FlowletFormer drops more substantially under limited data, suggesting that the traffic patterns in this dataset are more complex and require more data to learn effectively.
\begin{table*}[ht]
\centering
\caption{Few-shot Analysis (F1-score) on ISCXVPN2016, ISCX-Tor2016, and CSTNET-TLS 1.3 datasets.}
\scriptsize
\renewcommand\arraystretch{1.1}
\setlength{\tabcolsep}{1.5pt}
\begin{tabular}{c|cccc|cccc|cccc|cccc}
\toprule
\textbf{Dataset} & \multicolumn{4}{c|}{ISCX-VPN(Service)} & \multicolumn{4}{c|}{Tor} & \multicolumn{4}{c|}{ISCX-VPN(App)} & \multicolumn{4}{c}{CSTNET-TLS} \\
\midrule
\textbf{Size}  & 100\% & 40\% & 20\% & 10\% & 100\% & 40\% & 20\% & 10\% & 100\% & 40\% & 20\% & 10\% & 100\% & 40\% & 20\% & 10\% \\
\midrule
AppScanner     & 0.8546 & 0.7512 & 0.6074 & 0.5065 & 0.7848 & 0.7456 & 0.6195 & 0.5401 & 0.6874 & 0.4382 & 0.5320 & 0.2222 & 0.7023 & 0.6416 & 0.5661 & 0.4018 \\
BIND           & 0.7699 & 0.6625 & 0.4603 & 0.3290 & 0.8439 & 0.7222 & 0.5582 & 0.5269 & 0.5609 & 0.3182 & 0.2003 & 0.1992 & 0.4189 & 0.3558 & 0.2933 & 0.2299 \\
CUMUL          & 0.6884 & 0.5244 & 0.3873 & 0.4511 & 0.6401 & 0.5749 & 0.5252 & 0.5775 & 0.4554 & 0.3081 & 0.2673 & 0.1550 & 0.5493 & 0.4598 & 0.3659 & 0.2982 \\
DF             & 0.3934 & 0.3349 & 0.2596 & 0.0686 & 0.5492 & 0.4850 & 0.2499 & 0.1557 & 0.2289 & 0.1906 & 0.1476 & 0.1234 & 0.4933 & 0.2428 & 0.0449 & 0.0543 \\
FSNet          & 0.9051 & 0.8384 & 0.7078 & 0.3931 & 0.6028 & 0.5426 & 0.4080 & 0.5743 & 0.4677 & 0.4795 & 0.4752 & 0.2738 & 0.7311 & 0.7132 & 0.6662 & 0.5946 \\
GraphDApp      & 0.7429 & 0.5713 & 0.6137 & 0.2762 & 0.6383 & 0.5780 & 0.4622 & 0.4895 & 0.4853 & 0.2427 & 0.2203 & 0.1944 & 0.6890 & 0.4948 & 0.4372 & 0.3303 \\
Beauty         & 0.5387 & 0.3635 & 0.4063 & 0.4797 & 0.2251 & 0.0676 & 0.0593 & 0.1571 & 0.2964 & 0.2841 & 0.2911 & 0.2748 & 0.2324 & 0.0799 & 0.1059 & 0.0848 \\
ET-BERT        & 0.8393 & 0.3980 & 0.2450 & 0.2583 & 0.7453 & 0.4959 & 0.3749 & 0.3512 & 0.7066 & 0.6465 & 0.5728 & 0.4631 & 0.7700 & 0.7039 & 0.6117 & 0.4819 \\
TrafficFormer  & 0.8821 & 0.6827 & 0.5595 & 0.3909 & 0.7405 & 0.4989 & 0.3506 & 0.3674 & 0.6962 & 0.6085 & 0.5404 & 0.4320 & 0.7704 & 0.7084 & 0.6277 & 0.5660 \\
YaTC           & 0.8279 & 0.0801 & 0.0721 & 0.0947 & 0.7472 & 0.6587 & 0.4994 & 0.0721 & 0.7254 & 0.6489 & 0.5939 & 0.1805 & 0.8140 & 0.7538 & 0.6375 & 0.5040 \\
FlowletFormer  & \textbf{0.9364} & \textbf{0.8956} & \textbf{0.7356} & \textbf{0.8106} & \textbf{0.9116} & \textbf{0.7829} & \textbf{0.7166} & \textbf{0.5917} & \textbf{0.7712} & \textbf{0.8009} & \textbf{0.6224} & \textbf{0.5813} & \textbf{0.8473} & \textbf{0.8171} & \textbf{0.7273} & \textbf{0.6249} \\
\bottomrule
\end{tabular}
\label{tab:fewshot1}
\end{table*}

\begin{table*}[ht]
\centering
\caption{Few-shot Analysis (F1-score) on Browser, USTC-TFC, CIC-IDS2017, and CIC-IoT2022 datasets.}
\scriptsize
\renewcommand\arraystretch{1.1}
\setlength{\tabcolsep}{1.5pt}
\begin{tabular}{c|cccc|cccc|cccc|cccc}
\toprule
\textbf{Dataset} & \multicolumn{4}{c|}{Browser} & \multicolumn{4}{c|}{USTC-TFC} & \multicolumn{4}{c|}{CICIDS2017} & \multicolumn{4}{c}{IoT} \\
\midrule
\textbf{Size}  & 100\% & 40\% & 20\% & 10\% & 100\% & 40\% & 20\% & 10\% & 100\% & 40\% & 20\% & 10\% & 100\% & 40\% & 20\% & 10\% \\
\midrule
AppScanner     & 0.5733 & 0.3756 & 0.3524 & 0.1838 & 0.8976 & 0.7407 & 0.6799 & 0.5733 & 0.8630 & 0.8158 & 0.7924 & 0.7265 & 0.8288 & 0.6925 & 0.5149 & 0.4027 \\
BIND           & 0.5738 & 0.5288 & 0.4229 & 0.1304 & 0.7115 & 0.6609 & 0.7008 & 0.5653 & 0.8788 & 0.8377 & 0.7673 & 0.6860 & 0.6435 & 0.6428 & 0.4828 & 0.3444 \\
CUMUL          & 0.5207 & 0.3986 & 0.3742 & 0.1500 & 0.5183 & 0.4654 & 0.3753 & 0.3631 & 0.6951 & 0.5602 & 0.5031 & 0.4991 & 0.6687 & 0.5582 & 0.5479 & 0.2113 \\
DF             & 0.1627 & 0.2068 & 0.1399 & 0.0833 & 0.3059 & 0.2949 & 0.1508 & 0.1507 & 0.5940 & 0.2704 & 0.2956 & 0.2007 & 0.1647 & 0.1019 & 0.0382 & 0.0096 \\
FSNet          & 0.6410 & 0.4364 & 0.4444 & 0.1852 & 0.8042 & 0.6406 & 0.5563 & 0.7091 & 0.8144 & 0.7558 & 0.7244 & 0.5827 & 0.7804 & 0.5518 & 0.6089 & 0.4857 \\
GraphDApp      & 0.4912 & 0.3238 & 0.2484 & 0.2875 & 0.8249 & 0.7729 & 0.6429 & 0.5219 & 0.8647 & 0.8266 & 0.6106 & 0.6531 & 0.6767 & 0.4627 & 0.3642 & 0.1766 \\
Beauty         & 0.1040 & 0.2456 & 0.1730 & 0.0965 & 0.3796 & 0.4208 & 0.4439 & 0.2026 & 0.6567 & 0.4464 & 0.3457 & 0.3357 & 0.0296 & 0.1290 & 0.0593 & 0.0965 \\
ET-BERT        & 0.3439 & 0.3616 & 0.2280 & 0.2500 & 0.9666 & 0.9669 & 0.9286 & 0.8950 & 0.8911 & 0.8764 & 0.7346 & 0.7405 & 0.8244 & 0.7349 & 0.5630 & 0.4338 \\
TrafficFormer  & 0.5154 & 0.1520 & 0.1645 & 0.1154 & 0.9707 & \textbf{0.9703} & 0.9406 & \textbf{0.9432} & 0.8785 & 0.8725 & 0.7622 & 0.6918 & 0.8048 & 0.7578 & 0.5437 & 0.5190 \\
YaTC           & 0.3320 & 0.4761 & 0.4176 & 0.1613 & \textbf{0.9746} & 0.9480 & \textbf{0.9655} & 0.9159 & 0.8959 & 0.8854 & 0.6714 & 0.5902 & 0.8288 & 0.7243 & 0.7665 & 0.0758 \\
FlowletFormer  & \textbf{0.6684} & \textbf{0.6230} & \textbf{0.6553} & \textbf{0.3095} & 0.9675 & 0.9553 & 0.9457 & 0.9380 & \textbf{0.9079} & \textbf{0.8997} & \textbf{0.8610} & \textbf{0.8510} & \textbf{0.8859} & \textbf{0.8237} & \textbf{0.8180} & \textbf{0.6152} \\\bottomrule
\end{tabular}
\label{tab:fewshot2}
\end{table*}

\subsection{More Clarification of Word Analogies Similarity Analysis}
\label{Appendix:WordAnalogies}
To further clarify the purpose and design of the \textbf{Word Analogies Similarity Analysis} in Section~4.6, we emphasize that this experiment is not a classification task, but rather a semantic probing analysis inspired by methodologies from natural language processing.

In NLP, analogical reasoning tasks (e.g., \textit{``king - man + woman $\approx$ queen''}) are commonly used to evaluate whether pretrained language models capture meaningful token relationships. Following this intuition, we designed an analogous probing task in the context of network traffic to examine the semantic structure of token embeddings learned during pretraining.

Specifically, we selected three well-known HTTP-related port numbers (\textbf{80}, \textbf{8080}, and \textbf{8000}) and analyzed their relative positions in the learned embedding space using cosine similarity. These ports are commonly used for HTTP services and frequently co-occur in real-world traffic, thus forming a semantically coherent group.

Our experimental results show that FlowletFormer captures the semantic similarity between these ports more accurately than baseline models. This suggests that the model has developed a deeper understanding of protocol-layer semantics and is capable of organizing related concepts (e.g., similar ports) in a meaningful embedding space.

\subsection{More Computational Cost and Complexity}
\label{Appendix:ComputationalCost}

Table~\ref{tab:computation} reports the full comparison of FlowletFormer against two baseline models (ET-BERT and TrafficFormer) across the three experimental phases: pretraining (6 × V100 GPUs, 200 K steps), fine-tuning (1 × V100 GPU, full epochs), and inference (throughput in samples/sec). All runs were carried out under identical hardware and configuration settings to ensure a fair evaluation of runtime, per-step/epoch granularity, and GPU memory usage.

\begin{table*}[ht]
\centering
\caption{Computational efficiency comparison across pretraining, fine-tuning, and inference.}
\label{tab:computation}
\scriptsize
\begin{tabular}{lcccccc}
\toprule
\textbf{Phase} & \textbf{Model} & \textbf{GPUs} & \textbf{Time} & \textbf{Unit/Granularity} & \textbf{GPU Memory (GB)} \\
\midrule
\multirow{3}{*}{Pretraining} 
  & FlowletFormer & 6 & 42 h  & 75.67 s / 100 steps & 28 \\
  & ET-BERT       & 6 & \textbf{41 h} & \textbf{73.87 s / 100 steps} & 28 \\
  & TrafficFormer & 6 & 45 h  & 82.00 s / 100 steps & 28 \\
\midrule
\multirow{3}{*}{Fine-tuning} 
  & FlowletFormer & 1 & \textbf{1,153 s} & \textbf{57.65 s / epoch}    & 17 \\
  & ET-BERT       & 1 & 1,177 s & 58.85 s / epoch           & 17 \\
  & TrafficFormer & 1 & 1,158 s & 57.90 s / epoch           & 17 \\
\midrule
\multirow{3}{*}{Inference} 
  & FlowletFormer & 1 & —      & 150.04 samples/sec         & —  \\
  & ET-BERT       & 1 & —      & \textbf{148.92 samples/sec} & —  \\
  & TrafficFormer & 1 & —      & 150.45 samples/sec         & —  \\
\bottomrule
\end{tabular}
\end{table*}

\subsection{Limitation}
\label{Appendix:limtation}
Though FlowletFormer achieves fine‐grained behavioral analysis within each flowlet, it still has several limitations.

First, the fixed maximum input length forces us to split long flows into shorter flowlets. While this enables detailed study of intra‐flow behaviors, it prevents the model from learning unified patterns over entire long flows, which may be crucial for detecting certain sophisticated or slow‐evolving anomalies.

Second, our Field Tokenization treats each protocol field as an independent ``word'' analogous to treating every single Chinese character as a separate token. Although this captures the finest‐grained units, it cannot model semantic entities that span multiple fields. In future work, we could adopt Chinese word segmentation techniques to merge common adjacent fields into higher‐level tokens

Third, because FlowletFormer is based on the BERT architecture, both pretraining and real-time inference demand substantial GPU resources. This high computational and memory overhead may limit deployment in resource-constrained environments or scenarios requiring very high throughput.

Lastly, despite introducing protocol‐stack alignment and field‐aware pretraining objectives, the internal decision process of FlowletFormer remains difficult to interpret and audit. This lack of transparency can be problematic in high-security settings where explainability and trust are paramount.

\paragraph{Broader Impacts}  
While FlowletFormer can significantly enhance the accuracy of anomaly detection and threat mitigation—thereby contributing to more secure and reliable networks—it also carries potential risks. On the positive side, better traffic classification aids in detecting malicious activities (e.g., DDoS, malware propagation) and supports privacy-preserving analytics by filtering out sensitive flows before further processing. On the negative side, the same techniques could be repurposed for intrusive traffic monitoring or profiling of users, raising privacy and ethical concerns. To mitigate such risks, we advocate for transparent deployment policies, strict access controls, and regular audits of model usage.